\documentclass[sigconf]{acmart}
\usepackage{multirow}
\usepackage{balance}
\usepackage{epstopdf}
\usepackage{graphicx}
\usepackage{makecell}
\usepackage{multirow}
\usepackage{array}
\usepackage{threeparttable}
\AtBeginDocument{%
  \providecommand\BibTeX{{%
    \normalfont B\kern-0.5em{\scshape i\kern-0.25em b}\kern-0.8em\TeX}}}

\renewcommand\footnotetextcopyrightpermission[1]{} % removes footnote with conference information in first column
\begin{document}
\fancyhead{}
\title{Learnable Privacy-Preserving Anonymization for Pedestrian Images}


% \author{Submission ID: 13}

% \author{Junwu Zhang$^1$, Mang Ye$^{1*}$, Yao Yang$^{2}$}
% \affiliation{%
%   $^1$\institution{National Engineering Research Center for Multimedia Software, Institute of Artificial Intelligence, \\ Hubei Key Laboratory of Multimedia and Network Communication Engineering, \\ School of Computer Science, Wuhan University, Wuhan, China}
%   $^2$\institution{Zhejiang Lab, Hangzhou, China}
% }
% \email{{junwuzhang, yemang}@whu.edu.cn, yangyao@zhejianglab.com}

%   \city{Hangzhou}
%   \country{China}
  
\author{Junwu Zhang}
\email{junwuzhang@whu.edu.cn}
\affiliation{%
  \institution{School of Computer Science, Wuhan University}
  \city{Wuhan}
  \country{China}
  }

\author{Mang Ye}
\email{yemang@whu.edu.cn}
\authornotemark[0]
\authornote{Corresponding author.}
\affiliation{%
  \institution{School of Computer Science, Wuhan University}
%   \institution{Hubei Luojia Laboratory}
  \city{Wuhan}
  \country{China}
  }

\author{Yao Yang}
\email{yangyao@zhejianglab.com}
\affiliation{%
  \institution{Zhejiang Lab}
  \city{Hangzhou}
  \country{China}
  }

\renewcommand{\shortauthors}{Junwu Zhang, Mang Ye, Yao Yang}

\begin{abstract}
This paper studies a novel privacy-preserving anonymization problem for pedestrian images, which preserves personal identity information (PII) for authorized models and prevents PII from being recognized by third parties. Conventional anonymization methods unavoidably cause semantic information loss, leading to limited data utility. Besides, existing learned anonymization techniques, while retaining various identity-irrelevant utilities, will change the pedestrian identity, and thus are unsuitable for training robust re-identification models. To explore the privacy-utility trade-off for pedestrian images, we propose a joint learning reversible anonymization framework, which can reversibly generate full-body anonymous images with little performance drop on person re-identification tasks. The core idea is that we adopt desensitized images generated by conventional methods as the initial privacy-preserving supervision and jointly train an anonymization encoder with a recovery decoder and an identity-invariant model. We further propose a progressive training strategy to improve the performance, which iteratively upgrades the initial anonymization supervision. Experiments further demonstrate the effectiveness of our anonymized pedestrian images for privacy protection, which boosts the re-identification performance while preserving privacy. Code is available at \url{https://github.com/whuzjw/privacy-reid}.
\end{abstract}




%   <concept>
%       <concept_id>10010147.10010178.10010224.10010245.10010254</concept_id>
%       <concept_desc>Computing methodologies~Reconstruction</concept_desc>
%       <concept_significance>500</concept_significance>
%       </concept>
%   <concept>
%       <concept_id>10010147.10010178.10010224.10010226.10010239</concept_id>
%       <concept_desc>Computing methodologies~3D imaging</concept_desc>
%       <concept_significance>500</concept_significance>
%       </concept>
%  </ccs2012>

% \ccsdesc[500]{Computing methodologies~Reconstruction}
% \ccsdesc[500]{Computing methodologies~3D imaging}



\maketitle

\section{Introduction}

\begin{figure}[t]
\centering
  \includegraphics[ width=8cm, height=4.5cm]{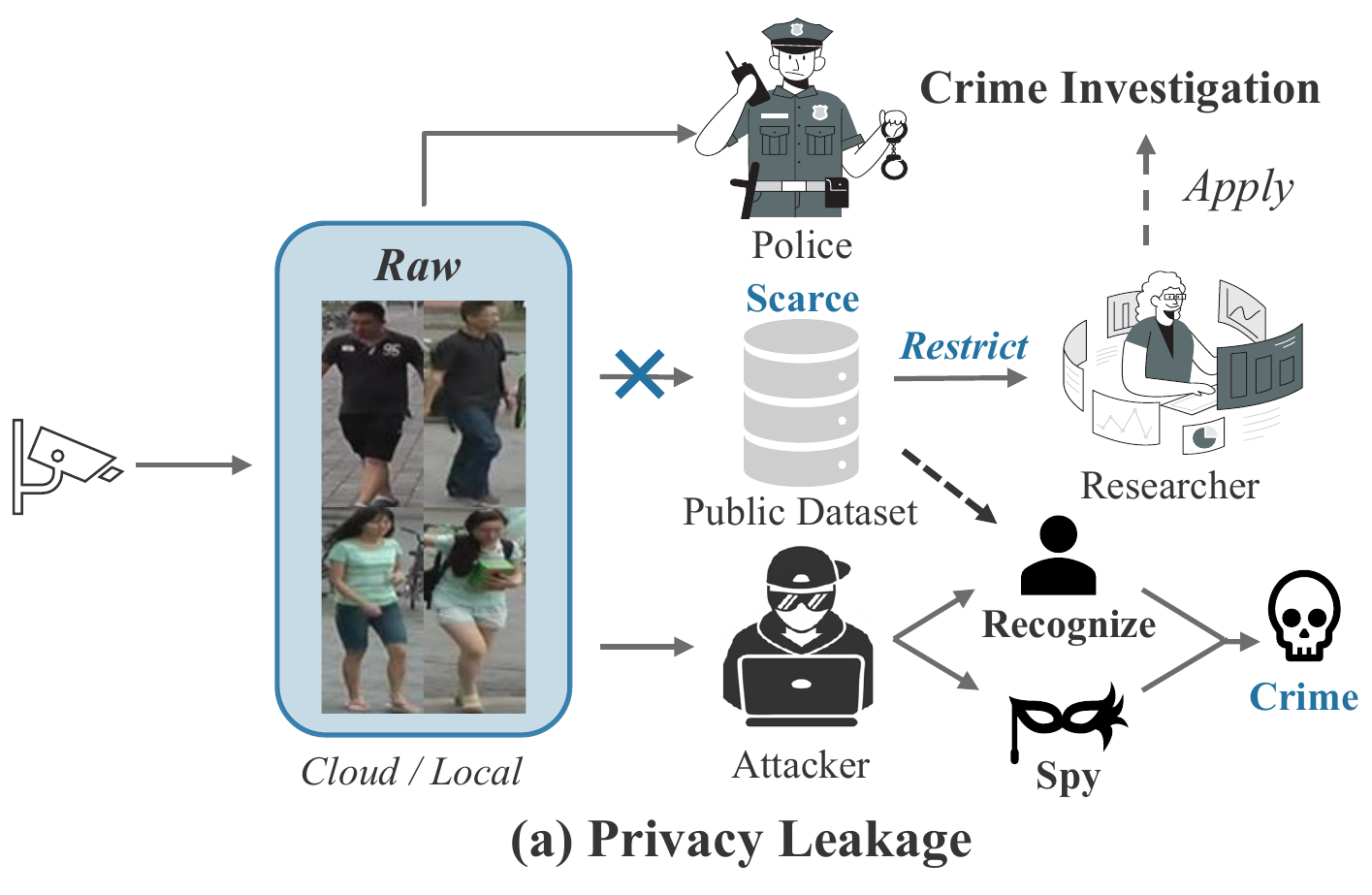}
  \includegraphics[ width=8cm, height=4.5cm]{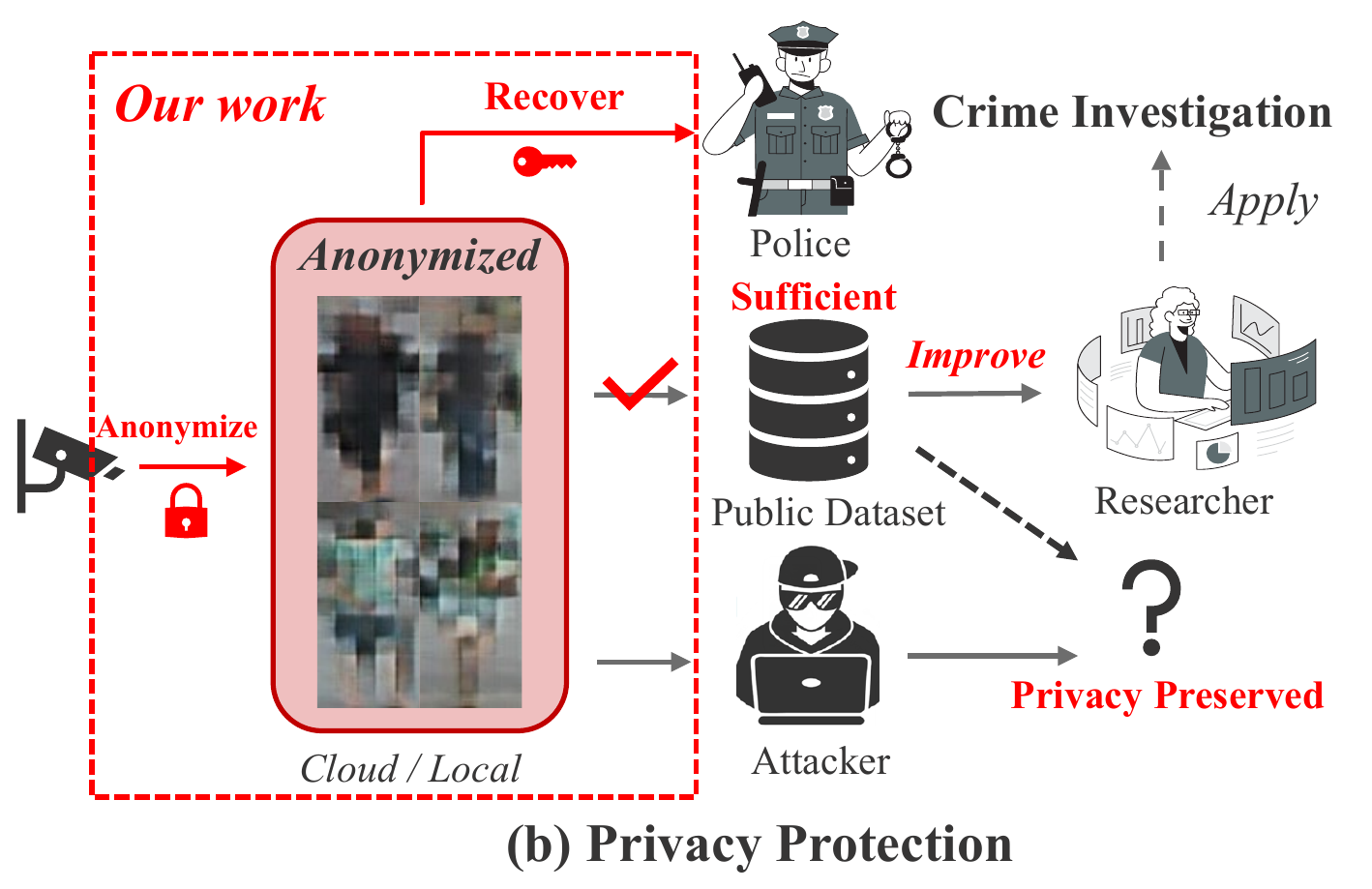}
  
  \caption{Motivation of our privacy-preserving system. (a) represents the risk of privacy leakage in current surveillance systems. The potential abuse of sensitive identity information leads to the deficiency of public datasets, which restricts related research. (b) represents our privacy protection method in surveillance systems. Our anonymized pedestrian images not only protect sensitive information from abuse, but also are suitable for person re-identification research that can be applied to crime investigation. Besides, original raw images cannot be recovered from our anonymized images by authorized users.}
  
\label{fig:motivation}
\end{figure}

\begin{figure*}[t]
\centering
  \includegraphics[ width=17cm, height=6.5cm]{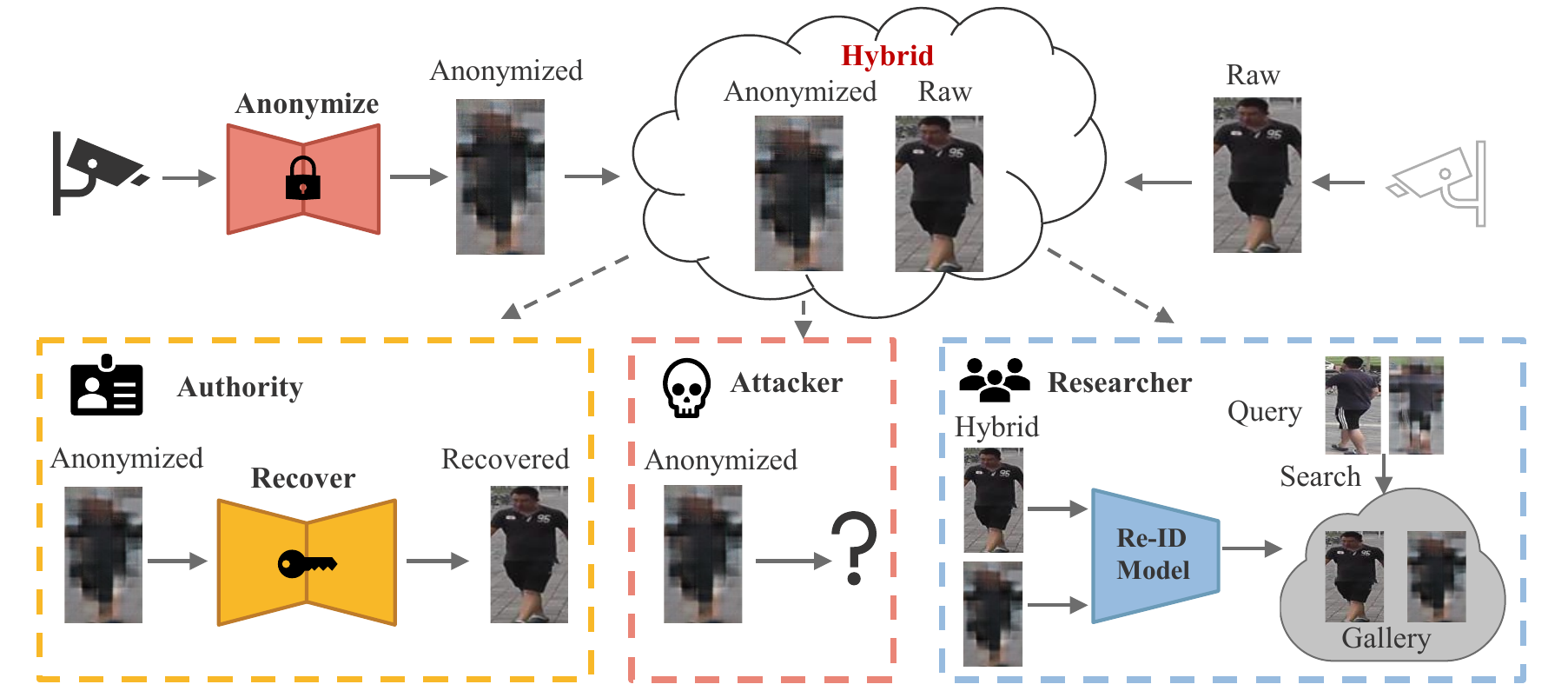}
  
  \caption{{Various utilities of our surveillance systems for different targets. Our anonymized images can not only protect privacy from abuse by malicious attackers, but also retain utilities for authority and the researchers. }}
  
\label{fig:application}
\end{figure*}

With the development of machine learning and the expansion of personal private data, various intelligent applications emerge and bring lots of utility value to individuals and the society. However, sensitive personal information raises serious privacy issues that are becoming increasingly prominent. Due to the privacy concerns, a lot of organizations have to make a compromise, e.g., Meta (Facebook) is shutting down its facial recognition software recently \cite{facebook}. Moreover, public research datasets are also influenced by the privacy concerns, e.g., DukeMTMC dataset \cite{ristani2016performance} and Tiny Image dataset \cite{torralba200880} were taken down and all the facial areas in ImageNet \cite{yang2021study} are blurred. As a result, privacy protection provides security while restricting technology improvement. Therefore, finding a privacy-utility trade-off point is of great importance.

In surveillance scenarios, privacy issues are particularly evident, as shown in Fig.~\ref{fig:motivation}(a). Ubiquitous surveillance systems take a huge number of raw pedestrian images and videos, which are stored in local storage or uploaded to the cloud servers. On one hand, it is useful for legal users in many scenarios, e.g., in crime investigation. On the other hand, this raises serious privacy concerns for individuals and public safety, since original images or videos contain sensitive information about the pedestrians, e.g., realistic identity information of a specific person or special community. Without careful protection, the highly sensitive information might be leaked and abused by malicious parties for nefarious purposes. For example, malicious attackers may recognize individual realistic identities and spy on the individuals for further crime or even forge them via Deepfake \cite{deepfake}. Moreover, due to potential privacy concerns, public surveillance datasets are scarce and sometimes taken down like DukeMTMC-reid \cite{ristani2016performance}. The deficiency of public datasets restricts the improvement of related research like person re-identification, and thus limits the development of intelligent video surveillance. Therefore, there is an urgent need to address the privacy issues for pedestrian images while retaining the utility value.

To tackle the above issues, a feasible anonymization solution is illustrated in Fig.~\ref{fig:motivation}(b). To prevent abuse after leakage, such an anonymization method is expected to have a satisfying visual obfuscation effect to ensure malicious parties cannot draw identity information from the anonymized images by human eyes. With the privacy preserved, the anonymized images can be securely used as public datasets. To retain the data utility for various users, the original raw images should be able to be recovered from the protected images for authorized utility, e.g., for police officers to investigate crime. Moreover, considering person re-identification (Re-ID) is imperative in intelligent surveillance systems with significant research impact and practical importance \cite{ye2021deep}, the anonymized images are supposed to retain necessary information for researchers to perform Re-ID tasks. In summary, \textit{an ideal anonymization method for pedestrian images should be reliable for privacy security, reversible for authorized utility and suitable for person re-identification research.}

Lots of research work has been done on image anonymization. \textit{Conventional anonymization methods}, e.g.,  blurring, pixelation and Gaussian noise adding, face the problem of semantic information loss, causing significant declines in utility value. To explore the privacy-utility trade-off, new techniques and mechanisms are proposed to de-identify images or videos to fool identification models while achieving various identity-irrelevant utility goals \cite{maximov2020ciagan, li2019anonymousnet, chen2021perceptual, ren2018learning, proencca2020uu, you2021reversible}, such as reversibility \cite{proencca2020uu}, privacy preserving action detection \cite{ren2018learning}, smile recognition \cite{you2021reversible} and so on. However, these anonymization techniques change the original individual identity to get a low identification rate by recognition model. In terms of crime investigation scenario, the identity variance is unsuitable for person re-identification task which is identity-relevant and requires that original and anonymized images of a specific pedestrian share the same virtual identity. Recently, Dietlmeier et al. \cite{dietlmeier2022improving} propose an anonymization dataset for Re-ID, which detect and blur the facial regions. However, non-face regions may also cause privacy leakage and the original images cannot be reconstructed from their anonymized images. 
%With the identifiable and biometric information concealed from human eyes, the Re-ID model can still perform well by capturing the insensitive features (e.g., the clothes color).  

To achieve the goal illustrated in Fig.~\ref{fig:motivation}(b), we propose a new reversible anonymization framework for pedestrian images, which can reversibly generate full-body anonymous images with little performance drop on Re-ID task. As shown in Fig.~\ref{fig:application}, the identity information of our anonymized pedestrian images can be invisible to attackers, but recoverable for authorized users and computable for researchers to perform Re-ID task on hybrid domains (raw and anonymized). The core idea of our work is that we first desensitize raw images by conventional methods (i.e., blurring, pixelation, or noise adding). Then these desensitized images are adopted as initial supervision images for an anonymization encoder which can translate raw images to privacy-preserving images in a learnable manner. To preserve necessary features for recovery and person re-identification, we jointly optimize the anonymization encoder with a recovery decoder and a Re-ID model. Through supervised and joint learning, our anonymized images can achieve good performance on privacy protection, recovery, and person re-identification. Besides, to further improve Re-ID performance, we propose a progressive training strategy referred to as \textit{supervision upgradation}. The supervision is upgraded by replacing the original desensitized images with the learned anonymized images, which are constrained by both privacy protection and Re-ID performance. Our main contributions are summarized as follows: 
\begin{itemize}
\item To the best of our knowledge, we are the first to explore the privacy-utility trade-off for pedestrian images from a Re-ID perspective, in which anonymized images cannot be recognized by third parties, but are recoverable for authorized users and suitable for person re-identification research.
\item We propose a reversible anonymization framework for Re-ID, which jointly optimizes an anonymization encoder with a recovery decoder and achieves the goal of obfuscating the image while keeping the identity for the authorized model.
\item We design a progressive training strategy called \textit{supervision upgradation}, which improves Re-ID performance by progressively upgrading the supervision of anonymization target.
\item We experimentally show that our anonymized images achieve good performance for privacy protection, recovery, and person re-identification tasks.
\end{itemize}

%-------------------------------------------------------------------------

\section{Related Work}

\textbf{Person Re-IDentification.}
Re-ID aims at retrieving a person of interest across multiple non-overlapping cameras \cite{ye2021deep}. 
% Early research mainly focused on handcrafted feature construction with body structures \cite{gray2008viewpoint, farenzena2010person, yang2014salient, liao2015person, matsukawa2016hierarchical} or distance metric learning \cite{zheng2011person, hirzer2012relaxed, koestinger2012large, xiong2014person, liao2015efficient, yu2018unsupervised}. 
With the development of deep neural networks, many works adopt deep convolutional neural networks (CNNs) as the backbone to extract the features of person images \cite{ye2021collaborative, ye2021dynamic, ye2022aug}, and incorporate domain generalization \cite{zhou2021domainsurvey, zhou2021domain} to generalize better to unseen domains \cite{zhou2019omni, zhou2021learning}. The CNN-based baselines \cite{wang2018learning, luo2019strong, ye2021deep}, such as AGW \cite{ye2021deep} and so on, achieve great success and play a key role in Re-ID community.
However, public Re-ID datasets face the challenge of privacy concerns, e.g., DukeMTMC-reID \cite{ristani2016performance} dataset was taken down due to privacy issues. To tackle this problem, we propose an anonymization method for Re-ID research, which can protect privacy while retaining necessary features for Re-ID tasks.

\textbf{Image-to-Image Translation.} Image-to-image translation is the task of transforming original images into the target images with a different style. Zhu \textit{et al.} proposed Pix2pix network \cite{isola2017image} and its unsupervised variant CycleGAN \cite{zhu2017unpaired} which achieved impressive performance on paired and unpaired cross-domain image translation. In this work, we use two Pix2pix networks for anonymization and recovery. Other advanced models can also be applied.

\textbf{Face Anonymization.}  Conventional face anonymization methods include pixelation, blurring, and noise adding. However, these methods cause semantic information loss, leading to performance degradation in detection and recognition. Therefore, researchers proposed many learnable anonymization methods based on face swapping \cite{li2019anonymousnet, ren2018learning, maximov2020ciagan, proencca2020uu, chen2021perceptual, sun2018natural, sun2018hybrid, gafni2019live, hukkelaas2019deepprivacy, kuang2021unnoticeable, kuang2021effective, wu2018privacy} to preserve important features for various utilities. 
% Among them, replacing the face with a synthetic face by GAN \cite{Goodfellow2014GenerativeAN} is a common method. Ren \textit{et al.} \cite{ren2018learning} proposed a pipeline to confuse both humans and machines in face identification while producing action detection. Li \textit{et al.} \cite{li2019anonymousnet} proposed a framework that can generate photo-realistic images with fake identities and is capable of balancing privacy and usability. Maximov \textit{et al.} proposed CIAGAN \cite{maximov2020ciagan} to have explicit control over generated face appearance and can be straightforwardly extended to full human bodies. UU-Net \cite{proencca2020uu} reversibly generates face-swapped person images for surveillance security. 
However, in some special cases, generated faces might overlap with realistic faces. Therefore, You \textit{et al.} \cite{you2021reversible} proposed a reversible face privacy-preserving framework based on a learned mosaic for smile recognition. Although the sole smile recognition task is easy and insufficient, the impressive results show that the semantic information for recovering and recognition can be invisibly embedded in protected images. This idea inspires us to anonymize pedestrian images for person re-identification.

\textbf{Privacy-Preserving Methods.} 
% To tackle the privacy concerns, a lot of approaches are proposed, including differential privacy \cite{dwork2006calibrating, dwork2008differential}, federated learning \cite{mcmahan2017communication, kairouz2019advances}, and so on. \textit{Differential privacy} is a technique that ensures the removal or addition of a single item does not (substantially) affect the outcome of any analysis \cite{dwork2008differential}. Instead of avoiding information leakage from the database, our method fundamentally protects privacy from the source of collecting data. \textit{Federated learning} is a learning technique that allows users to collectively reap the benefits of shared models trained from this rich data, without the need to centrally store it \cite{mcmahan2017communication}. By contrast, our method protects privacy from the source, and thus the data collected from users and public cameras can securely be uploaded to the cloud and centrally stored for easy access.
To tackle the privacy concerns, a lot of approaches are proposed, including differential privacy \cite{dwork2006calibrating, dwork2008differential}, federated learning \cite{mcmahan2017communication, kairouz2019advances}, and so on.
By contrast, our method protects privacy from the source, and thus the protected data can be stored securely and centrally for easy access.

%-------------------------------------------------------------------------
\begin{figure*}[t]
\centering
  \includegraphics[ width=17cm, height=5.5cm]{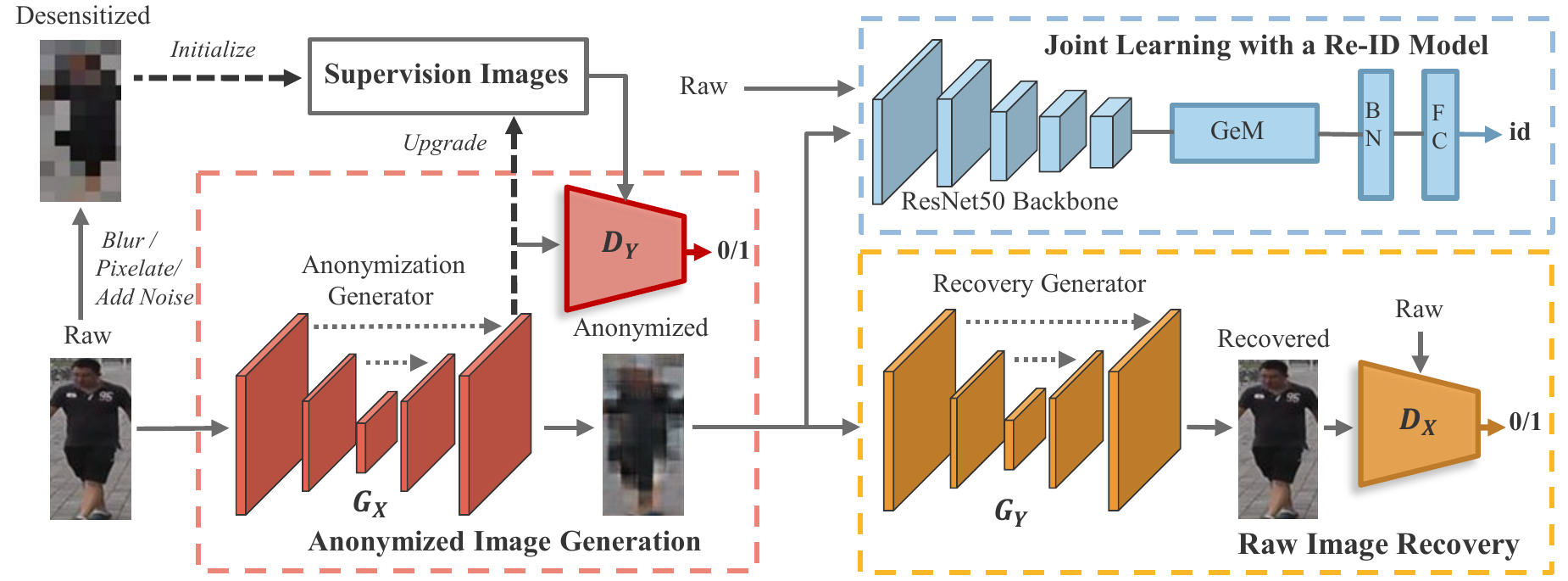}
  
  \caption{{Framework of the proposed method. The framework consists of four components. The anonymization model aims to produce privacy-preserving images in a learnable manner. The recovery model and Re-ID model are added to jointly optimize the anonymization model. The supervision (desensitized images) are progressively upgraded by learned anonymized images to further improve Re-ID performance. Different colors are corresponding to the different utilities illustrated in Fig.~\ref{fig:application}.}}
  
\label{fig:framework}
\end{figure*}

\section{Proposed Method}\label{sec:method}

In this section, we detail the methodology to reversibly anonymize images for Re-ID. As illustrated in Fig.~\ref{fig:framework}, our framework contains the following four components.
 1) \textit{Anonymized Image Generation $\S$~\ref{sec:ano}.} We exploit the power of image-to-image translation with conditional adversarial networks to generate anonymized images in a learnable manner. With learnable anonymization, further objectives can be achieved by joint learning.
 2) \textit{Raw Image Recovery $\S$~\ref{sec:rec}.} To achieve reversibility, a recovery model is added to embed necessary recovery information into the anonymized images.
 3) \textit{Joint Learning with a Re-ID Model $\S$~\ref{sec:reid}.} To preserve features for re-identification, we jointly optimize the anonymization generator with a Re-ID model. 
 4) \textit{Progressive Supervision Upgradation $\S$~\ref{sec:upgrade}.} In order to further improve Re-ID performance, the supervision images are progressively upgraded according to the performance of anonymized images on privacy protection and Re-ID.

\subsection {Anonymized Image Generation}\label{sec:ano}
% In order to generate anonymized images, our inspiration comes from the method of image-to-image translation, which can transform input images to the specified output images in a learnable manner. Our goal is to adopt blurred, pixelated or noise-added images as initial supervision to guide the generation of privacy-preserving images. Specifically, the pix2pix \cite{isola2017image} framework for image translation based on GAN \cite{Goodfellow2014GenerativeAN} is introduced, which contains two models, a U-Net128 \cite{ronneberger2015u} based anonymization generator and a ResNet50 \cite{he2016deep} based discriminator.

To generate anonymized images, our inspiration comes from the method of image-to-image translation, which can transform an input image into an output image of a specified form in a learnable manner. Our goal is to convert the original input image into a protected image form, and blurred, pixelated, or noise-added images can be adopted as initial supervision to guide the generation of privacy-preserving images. Specifically, the pix2pix \cite{isola2017image} framework for image translation based on GAN \cite{Goodfellow2014GenerativeAN} is introduced.

The training samples contain $\{x_i\}_{i=1}^n \in X$ with labels $\{label_i\}_{i=1}^m$, where $X$ are original images. We denote $\{y_i\}_{i=1}^n \in Y$, where $Y$ are supervision images which are initialized by conventionally desensitized images and can be further updated.
% Let $x$ and $y$ denote the data distribution $x \sim p_{\text {data }}(x)$ and $y \sim p_{\text {data }}(y)$. 
The goal of the anonymization generator $G_X$ is to learn a mapping function $G : X \rightarrow Y$. To achieve this goal, the generator $G_X$ is trained in an adversarial manner with a discriminator $D_Y$, where $G_X$ tries to generate images $G_X(x)$ similar to $y$ while $D_Y$ aims to distinguish between $G_X(x)$ and $y$. The adversarial objective of $G_X$ can be expressed as 
\begin{equation}
\begin{aligned}
\mathcal{L}_{adv_1} = \ & \frac{1}{n} \sum\nolimits_{i=1}^{n} \log (D_Y(x_{i}, y_{i})) \ + \\ 
& \frac{1}{n} \sum\nolimits_{i=1}^{n} \log (1-D_Y(x_{i}, G_X(x_{i}))),
\end{aligned}
\label{eq:adv1}
\end{equation}
where n represents the number of training samples within each batch and $G_X$ tries to minimize the objective against an adversarial $D_Y$ that tries to maximize it.

Besides, $L_1$ loss between $G_X(X)$ and $Y$ is adopted to guarantee that the learned function can map an individual input $x_{i}$ to a desired output $y_{i}$ \cite{isola2017image}.

\begin{equation}
\label{eq:la}
\mathcal{L}_{1_{ano}} = \frac{1}{n} \sum\nolimits_{i=1}^{n} \|y_{i}-G_X(x_{i})\|_{1}.
\end{equation}
The total loss of the anonymization generator is 
\begin{equation}
\label{eq:la}
\mathcal{L}_{ano} = \mathcal{L}_{adv_1} + \lambda_{L_1} \mathcal{L}_{1_{ano}},
\end{equation}
where $\lambda_{L_1}$ is a hyperparameter to reduce the artifacts \cite{isola2017image}.

\subsection{Raw Image Recovery}\label{sec:rec}

% The original raw images recovery can be regarded as the similar task of the anonymized images generation. To realize reversibility, the pix2pix \cite{isola2017image} framework can be used to jointly optimize the anonymization generator.
To make the process of generating anonymized images reversible, we design the raw image recovery to obtain generative raw images by inputting anonymized images. The pix2pix \cite{isola2017image} framework can be used to jointly optimize the anonymization generator. 

The raw image recovery is similar to anonymization process. In contrast to $G_X$, the recovery generator $G_Y$ is trained to learn a mapping function $F : Y \rightarrow X$ and produce recovered images $G_Y(G_X(x))$, which cannot be distinguished from original raw images $x$. The adversarial objective function is similar to Eq.~\ref{eq:adv1}:
\begin{equation}
\begin{aligned}
\mathcal{L}_{adv_2} \ = \ & \frac{1}{n} \sum\nolimits_{i=1}^{n} \log (D_X(G_X(x_{i}), x_{i})) \ + \\  & \frac{1}{n} \sum\nolimits_{i=1}^{n} \log (1-D_X(G_X(x_{i}), G_Y(G_X(x_{i})))),
\end{aligned}
\end{equation}
where $G_Y$ and $D_X$ oppose each other like $G_X$ and $D_Y$.

To make $G$ and $F$ forward cycle-consistent, i.e., $x \rightarrow G(x) \rightarrow F(G(x)) \approx x$, a cycle consistency loss \cite{zhu2017unpaired} is adopted: 
\begin{equation}
\mathcal{L}_{1_{rec}} = \frac{1}{n} \sum\nolimits_{i=1}^{n}  \|x_{i}-G_Y(G_X(x_{i}))\|_{1}.
\end{equation}

The total loss of the recovery generator is:
\begin{equation}
\label{eq:la}
\mathcal{L}_{rec} = \mathcal{L}_{adv_2} + \lambda_{L_1} \mathcal{L}_{1_{rec}}.
\end{equation}

\subsection{Joint Learning with a Re-ID Model} \label{sec:reid}
% In the field of person Re-ID, anonymization is a solution to the privacy issues. However, the desensitized images by conventional obfuscation methods cannot achieve satisfying Re-ID performance. For this reason, we propose to apply hybrid images (original and anonymized) to jointly train the Re-ID model and the anonymization generator. As a powerful baseline in the existing Re-ID research, AGW \cite{ye2021deep} is embedded in our architecture for joint learning. 
We design to embed the Re-ID model into our architecture for joint learning, which follows the powerful baseline AGW \cite{ye2021deep} in the existing Re-ID research. In the field of Re-ID, anonymization is a solution to the privacy issues. However, directly adopting desensitized images with conventional obfuscation methods will greatly affect the performance of Re-ID. Therefore, we propose to apply hybrid images (original and anonymized) to jointly train the Re-ID model and the anonymization generator. 

The Re-ID model takes paired inputs: raw images and their corresponding anonymized images with the same labels. By training on paired data, the Re-ID model learns to map original raw images and anonymized images of a specific person to the same virtual identity.
In detail, the Re-ID model contains three main components. \textit{a) backbone.} ResNet50 \cite{he2016deep} pre-trained on ImageNet \cite{deng2009imagenet} is adopted as the backbone with the stride of the last spatial down-sampling operation changed from 2 to 1. \textit{b) Generalized-mean (GeM) pooling.} The Global Average Pooling in the original ResNet50 is replaced with GeM \cite{ye2021deep} whose output is adopted for computing center loss and triplet loss during training process. \textit{c) BNNeck.} BNNeck \cite{luo2019strong} is added as a BN layer between features and FC layers.

% In AGW loss function $\mathcal{L}_{AGW}(x)$, three types of loss are combined for optimization, including identity classification loss ($\mathcal{L}_{id}$), center loss ($\mathcal{L}_{ct}$) \cite{wen2016discriminative} and weighted regularization triplet loss ($\mathcal{L}_{wrt}$) \cite{ye2021deep}. To make our Re-ID model adaptive to both raw and privacy-preserving scenarios, both raw and anonymized images are added as the input. Therefore the total loss of the Re-ID model is
The loss function is denoted by $\mathcal{L}_{AGW}(x)$, which combines three commonly used losses in Re-ID tasks including identity classification loss ($\mathcal{L}_{id}$), center loss ($\mathcal{L}_{ct}$) \cite{wen2016discriminative} and weighted regularization triplet loss ($\mathcal{L}_{wrt}$) \cite{ye2021deep} for optimization. To make our Re-ID model adaptive to both raw and privacy-preserving scenarios, both raw and anonymized images are added as the input. Therefore, the total loss of the Re-ID model is
\begin{equation}
\label{eq:la}
\mathcal{L}_{reid} = \mathcal{L}_{AGW}(x) + \mathcal{L}_{AGW}(G_X(x)).
\end{equation}

In summary, the final objective of our anonymization model for Re-ID on hybrid images is:
\begin{equation}
\mathcal{L} = \mathcal{L}_{ano} + \mathcal{L}_{rec} + \mathcal{L}_{reid} .
\end{equation}

\begin{figure}[t]
\centering
  \includegraphics[ width=8cm]{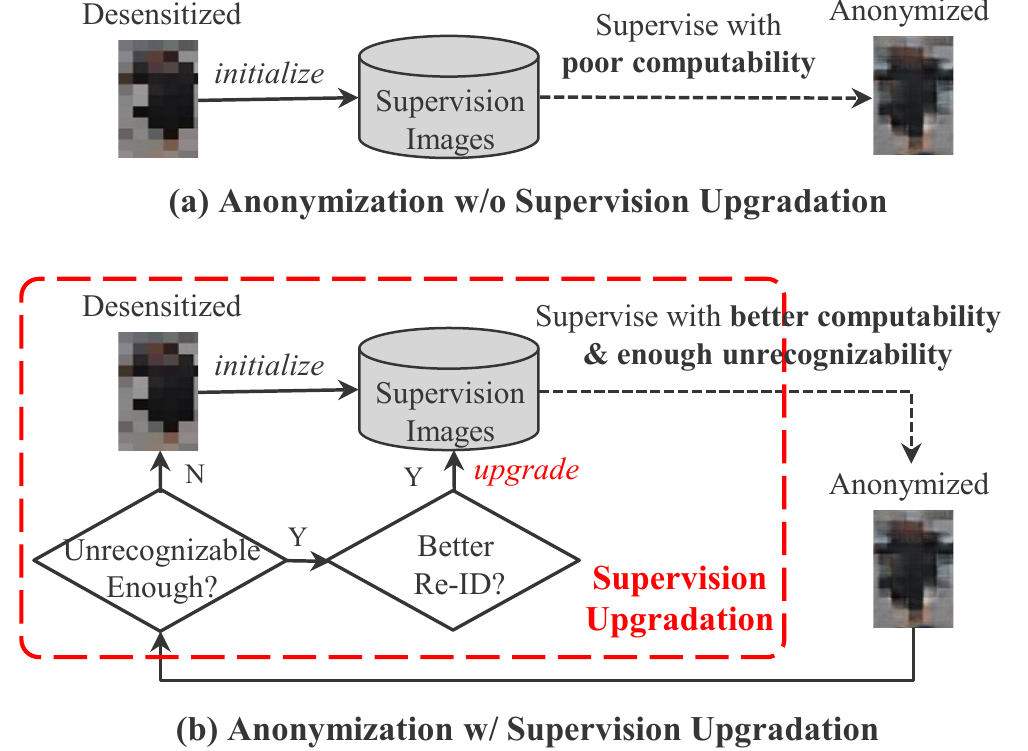}
  
%   \caption{{Illustration of supervision upgradation. “ano" represents our anonymized images while “des" indicates the desensitized images. $R1_{raw-ano}$ means the rank-1 accuracy with raw images as $query$ and our anonymized images as $gallery$. All three values of $\epsilon$ are small.}}
\caption{{Illustration of supervision upgradation. The upgradation is based on the performance of protection and Re-ID.}}
  
\label{fig:upgrade-explanation}
\end{figure}

\subsection{Progressive Supervision Upgradation}\label{sec:upgrade}
In $\S$~\ref{sec:reid}, a Re-ID model is added for joint learning to make anonymized images $G_X(x)$ suitable for identity preserving. However, initial supervision images (i.e., blurred, pixelated or noise-added images) are not optimal as final supervision images because the semantic information loss leads to identity variance, and thus restricts the improvement of Re-ID performance. Therefore, as shown in Fig.\ref{fig:framework} briefly and Fig.~\ref{fig:upgrade-explanation} in detail, we progressively upgrade the supervision images during the training process to satisfy both privacy protection (unrecognizability) and Re-ID constraints. Specifically, the privacy constraint is met when PSNR and SSIM calculated between anonymized images and raw images (i.e., $PSNR_{ano}$ and $SSIM_{ano}$) are lower than those calculated between desensitized images and raw images (i.e., $PSNR_{des}$ and $SSIM_{des}$) plus a small positive value (i.e., $\epsilon_{psnr}$ and $\epsilon_{ssim}$). The Re-ID constraint is met when the rank-1 accuracy (i.e., $R1_{raw-ano}$) with raw images as $query$ and anonymized images as $gallery$ is higher than the previous maximum rank-1 value. 
To begin with, we train a Re-ID model with raw and desensitized images as input and get the rank-1 value (i.e., $R1_{raw-des}$) with raw images as $query$ and desensitized images as $gallery$. Then, the desensitized images are adopted as the initial supervision images and the maximum rank-1 value is initialized to a $R1_{raw-des}$ plus small negative value ($-\epsilon_{r1}$).
While training, the supervision images will keep as desensitized images to guarantee the unrecognizability if the privacy need is not met and will be upgraded to $G_X(x)$ only when both constraints are satisfied. Through the supervision upgradation, our supervision images are becoming more adequate for Re-ID research while preserving privacy.

\begin{figure}[t]
\centering
  \includegraphics[ width=8cm]{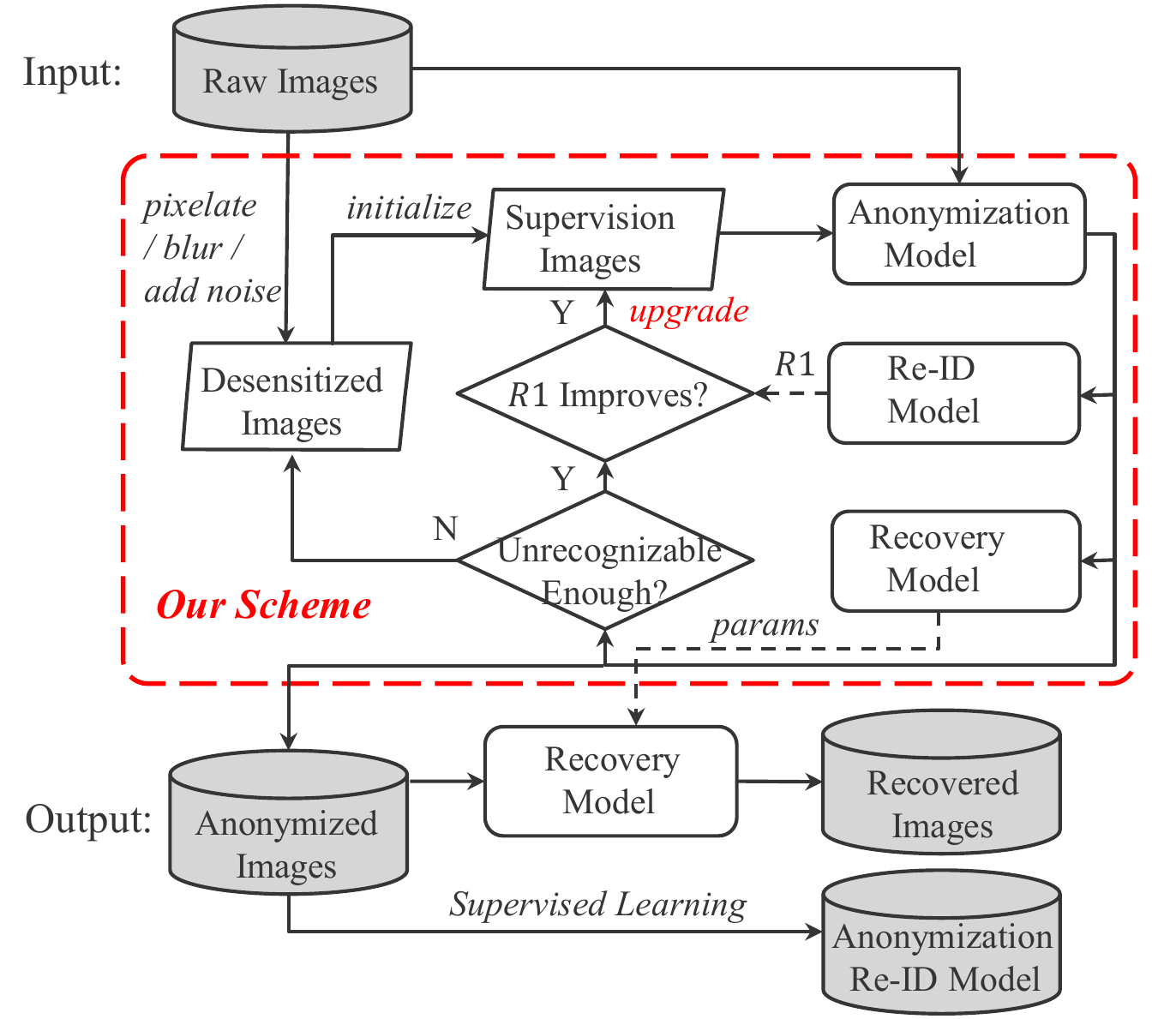}
  \put(-34.5,146.5){\scriptsize\textcolor{red}{ $\S$~\ref{sec:ano}}}
  \put(-34.5,83){\scriptsize\textcolor{red}{ $\S$~\ref{sec:rec}}}
  \put(-34.5,119){\scriptsize\textcolor{red}{ $\S$~\ref{sec:reid}}}
  \put(-102,145){\scriptsize\textcolor{red}{$\S$~\ref{sec:upgrade}}}
  
  \caption{{Flowchart of our training process. ``R1'' represents rank-1 accuracy with anonymized images as $query$ and desensitized images as $gallary$.}}
  
\label{fig:architecture}
\end{figure}
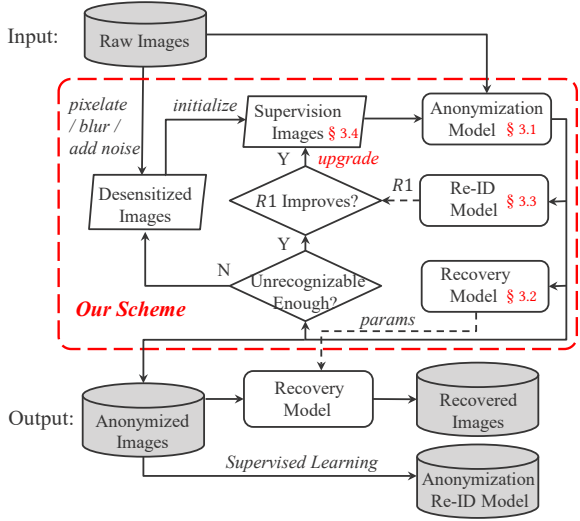

\subsection{Scheme of Training Process}\label{sec:scheme}
In Fig.~\ref{fig:architecture}, we show the flowchart of the training process. The input raw images are first desensitized by conventional methods, which initialize the supervision images. Under the supervision, the anonymization model learns to translate raw images to anonymized images. The anonymized images are then fed into the Re-ID model and recovery model to jointly learn to preserve necessary features for recovery and retrieval. Then our training process is continued with the supervision images upgraded according to the performance of our anonymized images on privacy protection and person re-identification. After training, the output anonymized images can be used to recover original raw images with the parameters of the trained recovery model. Meanwhile, proper Re-ID performance can be achieved after supervised learning on these anonymized images.

%-------------------------------------------------------------------------
\begin{figure*}[t]
\centering
  \includegraphics[ width=17cm, height=3.5cm]{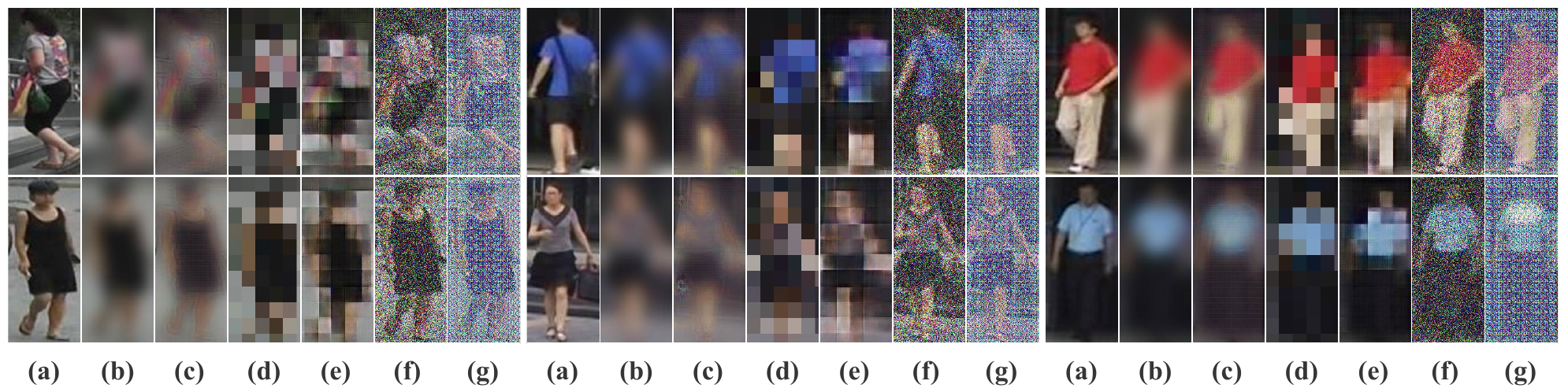}
  
  \caption{{Qualitative comparison on privacy protection. (a) raw images; (b)/(d)/(f) blurred/pixelated/noise-added images; (c)/(e)/(g) anonymized images guided by blurring/pixelation/noise adding. Best view in color. Zoom in for details.}}
  
\label{fig:qualitative}
\end{figure*}

%-------------------------------------------------------------------------

\section{Experimental Results}

\textbf{Preliminary.} In the following parts, we denote ``OI'' as the original raw images, ``PI'' as the protected images, ``w/ U'' and ``w/o U'' as anonymization with/without using supervision upgradation.

\subsection{Datasets and Evaluation Metrics}
We conduct experiments on three widely used datasets: Market-1501 \cite{zheng2015scalable}, MSMT17 \cite{wei2018person} and CUHK03 \cite{wei2018person}. The Market-1501 dataset comprises 32,668 annotated bounding boxes under six cameras. The MSMT17 dataset consists of 4,101 identities and 126,441 bounding boxes taken by a 15-camera network. The CUHK03 dataset contains 1,467 identities and 14,097 detected bounding boxes.
% \begin{itemize}
% \item \textbf{Market-1501} \cite{zheng2015scalable} comprises 32,668 annotated bounding boxes under six cameras. The training set has 12,936 images from 751 identities, while the testing set has 19,732 images from 750 identities. In testing, 3,368 images from 750 identities are used as queries to retrieve the matching persons in the gallery.

% \item \textbf{MSMT17} \cite{wei2018person} consists of 4,101 identities and 126,441 bounding boxes taken by an 15-camera network. The training set has 32,621 images from 1041 identities, and the testing set has 82,161 images from other 3060 identities.

% \item \textbf{CHUK03} \cite{wei2018person} contains of 1,467 identities and 14,097 automatically detected bounding boxes. The training set has 7,365 images from 767 identities, and the testing set has 5,332 images from 700 identities.
% 
% \end{itemize}
% 

We evaluate our model under image quality and re-identification metrics. For privacy protection and recovery, we adopt two widely used metrics: PSNR and SSIM \cite{wang2004image}. For Re-ID performance, Cumulative Matching Characteristics (\textit{a.k.a.,} Rank-k matching accuracy) \cite{wang2007shape}, mean Average Precision (mAP) \cite{zheng2015scalable}, and a new metric mean inverse negative penalty (mINP) \cite{ye2021deep} are used in our experiments.

 \subsection{Implementation Details}
\textbf{Training setup.} We first split the original training set into a new training set and a validation set in a ratio of $4:1$. Then we further split the validation set into a gallery set and a query set in the same ratio. All the performance while training is obtained by testing on the validation set. In all experiments, we jointly trained our three models for 120 epochs with batch size 64. All input images are resized to $256\times128$ and then desensitized by $blurring\ 12\times 12$, or $pixelation\ 24\times 24$, or adding $Gaussian\ noise\ N(0,0.5)$. We use Adam optimizer \cite{kingma2014adam} with $\beta_1 = 0.5, \beta_2 = 0.999$ for two generators and with default values $\beta_1 = 0.9, \beta_2 = 0.999$ for the Re-ID model. Learning rate is linearly increasing from 3.5 × 10$^{-5}$ to 3.5×10$^{-4}$ in the first 10 epochs, and then is decayed to 3.5×10$^{-5}$ and 3.5 × 10$^{-6}$ at 40th epoch and 80th epoch respectively. 

\textbf{Models of anonymization, recovery and Re-ID.} Our implementation for the anonymization and recovery models follow Pix2pix network \cite{isola2017image} and $\lambda_{L_1}$ is set to 100 as suggested by \cite{isola2017image}, while the Re-ID model follows the practice in \cite{ye2021deep} and uses the same hyperparameters. These might be replaced by other advanced methods.

\textbf{Supervision Upgradation.} To get a competitive effect of privacy protection, we set $\epsilon_{psnr}$ and $\epsilon_{ssim}$ (see in Fig.~\ref{fig:upgrade-explanation}) to small values of 1.0 and 0.05. And $\epsilon_{r1}$ is set to a small value of 0.05.

\subsection{Results of Privacy Protection}
% We show our privacy protection performance qualitatively and quantitatively. Besides, we perform human study to further show the effectiveness of our method on privacy protection. 
% Therefore, our anonymized images can be published as public datasets without privacy concerns.
%  Lower values of PSNR and SSIM indicate worse image quality, which indicates better privavy protection performance to some extent. Our anonymized images achieve similar PSNR and SSIM, compared to blurred and pixelated images. Being aware that a smaller or larger value of PSNR and SSIM does not necessarily imply higher or lower privacy protection extent \cite{zhang2018unreasonable}, we list these results only for reference purpose.

\textbf{Qualitative Results.} In Fig.~\ref{fig:qualitative}, a qualitative comparison of privacy protection performance is conducted. Compared to raw images, our anonymized images achieve good visual privacy protection performance. The individual's body contour line and details of face and clothes are all concealed, and thus one cannot obtain the identity from the anonymized images by human eyes. Compared with the desensitized images, our corresponding anonymized images attain a competitive visual obfuscation effect in a different style. 

\textbf{Quantitative Results.} Table~\ref{tab:privacy_reid} shows the Re-ID performance of unprotected AGW model on protected images. Compared with raw images (i.e., Raw), our anonymized images obtain extremely low rank-1 values and mAP values, showcasing that the common Re-ID model is not able to correctly identify our anonymized images. Compared with baselines, our anonymized images achieve similar Re-ID performance, indicating our anonymization method can obtain close privacy protection performance.

\begin{table}[t]
\caption{\label{tab:privacy_reid}{Re-ID performance of common AGW on protected images. ``Base'' means traditionally desensitized images. }}

 \begin{threeparttable}
 \resizebox{8cm}{22mm}{
\begin{tabular}{c|cc|cc|cc}
\hline
\multicolumn{1}{c|}{Dataset} &\multicolumn{2}{c|}{Market-1501}  & \multicolumn{2}{c|}{MSMT17} &\multicolumn{2}{c}{CUHK03}\\\hline
\multicolumn{1}{c|}{Images}  &  rank-1   & mAP   & rank-1   & mAP& rank-1   & mAP  \\\hline
\multicolumn{7}{l}{\textit{(a) Evaluation of blurring.}} \\\hline
\multirow{1}{*}{{ Base}} 
 & 20.6 & 8.7& 3.9 & 1.4& 1.9 & 2.5 \\\hline
\multirow{1}{*}{{ Ours}} 
 & 18.4 & 7.6& 8.3 & 2.6& 3.9 & 3.9 \\\hline
\multicolumn{7}{l}{\textit{(b) Evaluation of pixelation.}} \\\hline
\multirow{1}{*}{{ Base}} 
 & 20.2 & 9.1& 1.8 & 0.7& 2.1 & 2.3 \\\hline
\multirow{1}{*}{{ Ours}} 
 & 17.5 & 7.7& 6.7 & 2.1& 1.3 & 1.6 \\\hline
\multicolumn{7}{l}{\textit{(c) Evaluation of Gaussian noise.}} \\\hline
\multirow{1}{*}{{Base}} 
 & 0.6 & 0.4& 0.2 & 0.1& 0.1 & 0.3 \\\hline
\multirow{1}{*}{{Ours}} 
 & 1.4 & 0.6& 0.4 & 0.1& 0.2 & 0.4 \\\hline
 Raw & 95.7 & 88.6& 68.6 & 49.8& 67.3 & 65.8 \\\hline
 \end{tabular}
 }
 \end{threeparttable}
  
\end{table}

\begin{table}[t]
\caption{\label{tab:privacy}{Human evaluation results. ``Base'' represents conventional anonymization methods. Privacy value(\%) denotes verification accuracy by human eyes. Lower privacy value indicates better privacy protection, while higher Re-ID rank-1 accuracy means better Re-ID performance.}}  
 \begin{threeparttable}
%\resizebox{8cm}{32mm}{
\begin{tabular}{c|cc|cc|c|c}
\hline
\multicolumn{1}{c|}{Image} &\multicolumn{2}{c|}{A}  & \multicolumn{2}{c|}{B} &\multicolumn{1}{c|}{\multirow{2}{*}{{ Privacy value $\downarrow$}}}&\multicolumn{1}{c}{\multirow{2}{*}{{Re-ID rank-1 $\uparrow$}}}\\
\cline{1-5}
\multicolumn{1}{c|}{Method}   & OI   & PI  &  OI   & PI &\multicolumn{1}{c|}{~}& \multicolumn{1}{c}{~} \\\hline
\multicolumn{6}{l}{\textit{(a) Evaluation of blurring.}} \\\hline
\multirow{2}{*}{{ Base}}
& $\checkmark$ & & &$\checkmark$& 79 & 40.1\\
& & $\checkmark$ & &$\checkmark$& 82& 67.3 \\\hline
\multirow{2}{*}{{ Ours}}
& $\checkmark$ & & &$\checkmark$& 83& 88.2\\
& & $\checkmark$ & &$\checkmark$& 82 & 89.2 \\\hline
\multicolumn{6}{l}{\textit{(b) Evaluation of pixelation.}} \\\hline
\multirow{2}{*}{{ Base}}
& $\checkmark$ & & &$\checkmark$& 71& 75.3 \\
& & $\checkmark$ & &$\checkmark$& 75& 64.3 \\\hline
\multirow{2}{*}{{ Ours}}
& $\checkmark$ & & &$\checkmark$& 71& 88.5\\
& & $\checkmark$ & &$\checkmark$& 64& 87.0 \\\hline
\multicolumn{6}{l}{\textit{(c) Evaluation of Gaussian noise.}} \\\hline
\multirow{2}{*}{{ Base}}
& $\checkmark$ & & &$\checkmark$& 84& 50.8 \\
& & $\checkmark$ & &$\checkmark$& 83& 68.7 \\\hline
\multirow{2}{*}{{ Ours}}
& $\checkmark$ & & &$\checkmark$& 88& 83.5\\
& & $\checkmark$ & &$\checkmark$& 84 & 91.2 \\\hline
Upper &$\checkmark$  & & $\checkmark$ & & 92 & 95.7\\\hline
 \end{tabular} 
 \end{threeparttable}
\end{table}

\textbf{Human Evaluation.} Table~\ref{tab:privacy} shows the human evaluation of privacy protection effects of our method and baselines (blurring, pixelation and noise adding). We randomly sampled a pair of images from the raw or privacy-preserving Market-1501 testing set and ask participants whether the pair corresponds to the same person. The image pairs were divided into 13 groups (i.e., the 13 rows in Table~\ref{tab:privacy}) according to the protection method. Each group sampled 100 images that are distributed equally to 10 participants. The optimal privacy effect is when the privacy value equals 50\%, which indicates random guessing. Compared with raw image pairs (i.e., Upper), the pairs with our anonymized image achieve a substantially lower privacy value and a slight decrease in Re-ID accuracy. Compared with baselines, our method obtains comparable verification accuracy (i.e., privacy value) by human eyes and significantly better Re-ID rank-1 accuracy by trained Re-ID models. 

\subsection{Results of Recovery}
% We show our recovery performance in both qualitative and quantitative manners.

\begin{figure}[t]
\centering
  \includegraphics[ width=8cm, height=1.8cm]{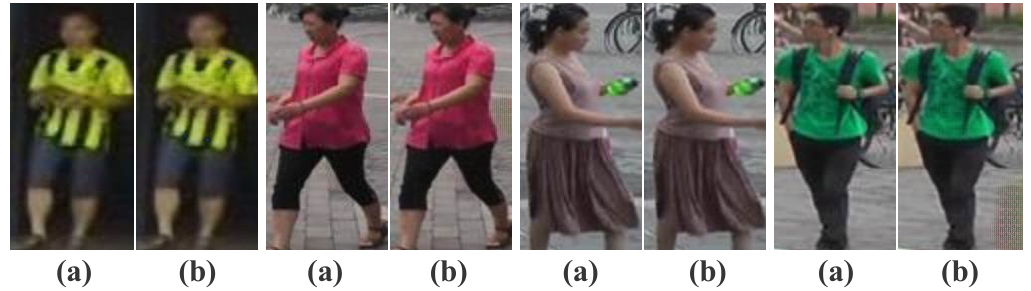}
    
  \caption{{Qualitative results of the recovered images. (a) represents raw images; (b) represents recovered images.}}
  
\label{fig:recovery}
\end{figure}

\textbf{Qualitative Results.} In Fig.~\ref{fig:recovery}, we qualitatively compare recovered images with raw images. Our recovered images achieve similar visual quality to original images. It is difficult to distinguish the recovered images from the original raw images by human eyes. 

\begin{table}[t]
 \caption{\label{tab:recovery}{Image quality of the recovered image. PSNR and SSIM are reported. Higher value indicates better quality.}}
 
 \begin{threeparttable}
%  \resizebox{8cm}{15.9mm}{
\begin{tabular}{c|cc|cc|cc}
\hline
\multicolumn{1}{c|}{Dataset} &\multicolumn{2}{c|}{Market-1501}  & \multicolumn{2}{c|}{MSMT17} &\multicolumn{2}{c}{CUHK03}\\\hline
\multicolumn{1}{c|}{Method}   & PSNR   & SSIM  &  PSNR   & SSIM   & PSNR   & SSIM  \\\hline
\multicolumn{7}{l}{\textit{(a) Evaluation of blurring.}} \\\hline
 \multirow{1}{*}{Ours} & 26.78 & 0.92  & 30.02 & 0.93  & 23.67 & 0.89 \\\hline
\multicolumn{7}{l}{\textit{(b) Evaluation of pixelation.}} \\\hline
 \multirow{1}{*}{Ours} & 29.74 & 0.93 &  25.94 & 0.89 & 27.00 & 0.93 \\\hline
\multicolumn{7}{l}{\textit{(c) Evaluation of Gaussian noise.}} \\\hline
 \multirow{1}{*}{Ours} & 26.80 & 0.92 & 27.80 & 0.92 & 23.00 & 0.91 \\\hline
 \multirow{1}{*}{Upper} & +$\infty$ & 1 & +$\infty$ & 1 & +$\infty$ & 1 \\\hline
% Ours  & - &  \bf{70.8} & \bf{84.1} & \bf{63.7} & \bf{39.6} & \bf{75.6} & \bf{90.1} & \bf{74.4} & \bf{67.8} \\\hline
 \end{tabular}
%  }
 \end{threeparttable}
   
\end{table}

\textbf{Quantitative Results.} In Table~\ref{tab:recovery}, we show the image quality of recovered images. Approximately, in all three datasets, the PSNR and SSIM values of recovered images are higher than 25 and 0.9, indicating that our recovered images have good image quality. Moreover, as shown in Table~\ref{tab:recovery_reid}, both the common AGW model and our protected Re-ID model can obtain Re-ID performance on our recovered images comparable to that of original raw images. Besides, compared to AGW model, our model suffers a slight degradation of performance on raw and recovered images since it is also trained to improve performance on the anonymized images whose style is obviously different from raw and recovered images.

\begin{table}[t]
 \caption{\label{tab:recovery_reid}{Re-ID performance of the recovered images. Rank at $r$ accuracy(\%) and mAP(\%) are reported. }}
 
 \begin{threeparttable}
 \resizebox{8cm}{12mm}{
\begin{tabular}{c|cc|cc|cc|cc}
\hline
\multicolumn{1}{c|}{} &\multicolumn{2}{c|}{images}&\multicolumn{2}{c|}{Market-1501}  & \multicolumn{2}{c|}{MSMT17} &\multicolumn{2}{c}{CUHK03}\\\hline
\multicolumn{1}{c|}{Model}   & OI & RI & r=1   & mAP  &  r=1   & mAP   & r=1   & mAP  \\\hline
\multirow{2}{*}{AGW} & \checkmark &  & 95.7 & 88.6 & 68.6 & 49.8& 67.3 & 65.8 \\
&  & \checkmark & 93.8 & 84.0 & 63.2 & 43.1& 64.6& 62.2 \\\hline
\multirow{2}{*}{Ours} & \checkmark &  & 91.7 & 78.2 & 48.6 & 29.8& 38.8 & 42.4 \\
&  & \checkmark & 90.4 & 75.5 & 49.8 & 29.5 & 33.3 & 33.1 \\\hline
 \end{tabular}
 }
 \end{threeparttable}
  
\end{table}

\subsection{Results of Person Re-identification}

\textbf{Experiments under Four Test Settings.} As shown in Table~\ref{tab:reid}, we test the Re-ID performance under four settings with different queries and galleries. These four settings represent different scenarios:  
1) \textbf{Original Setting} ($query=OI$, $gallery=OI$): This setting represents that we use original raw images for both query and gallery sets. The result shows that our proposed model achieves comparable performance to the existing widely used setting (Rank-1:91.7\% \textit{v.s.} 95.7\% on the Market1501 dataset) with only a minor performance drop. This demonstrates that the model trained on our anonymized dataset can still be applied to practical scenarios when the testing pedestrian images are not anonymized. It brings in another interesting research topic, i.e., designing algorithms on the anonymized dataset without the invasion of privacy, and testing in practical non-anonymized scenarios, which alleviates the major ethical concern of recent research on human subjects. 
2) \textbf{Protected Setting} ($query=PI$, $gallery=PI$): This setting indicates that we use privacy-preserving images for both query and gallery sets. Compared to baselines of blurring, pixelation and noise adding, our model achieves an average improvement of 26.8\%, 26.3\% and 28.0\% in Rank-1 on three datasets. Compared to the original setting, our model under protected setting achieves comparable results. The results indicate that our anonymized images are suitable for Re-ID research and can be applied to practical scenarios when the testing pedestrian images are anonymized. 
3) \textbf{Crossed Settings} ($query=OI$, $gallery=PI$ and $query=PI$, $gallery=OI$): These settings represent that we use different types of images for query and gallery sets. Compared to the baselines, our model significantly improves the performance on three datasets averagely by 34.2\% and 41.1\% for blurring, 22.4\% and 21.7\% for pixelation and 29.2\% and 33.0\% for noise adding. This indicates that our anonymization model is robust against privacy protection on the query and gallery sets. Besides, the performance is also comparable to the original setting, indicating that our model can be applied to Re-ID on hybrid images. It can be inferred that the feature distribution of our anonymized images has a close distance to that of raw images. Additionally, our model performs averagely better in the protected setting than in the crossed settings, probably because there still exists a minor domain gap between raw images and anonymized images. In summary, our anonymized images are suitable for Re-ID and the method can be applied to scenarios when testing images containing both raw and privacy-preserving images.

\begin{figure}[t]
\centering
  \includegraphics[ width=8cm, height=2.7cm]{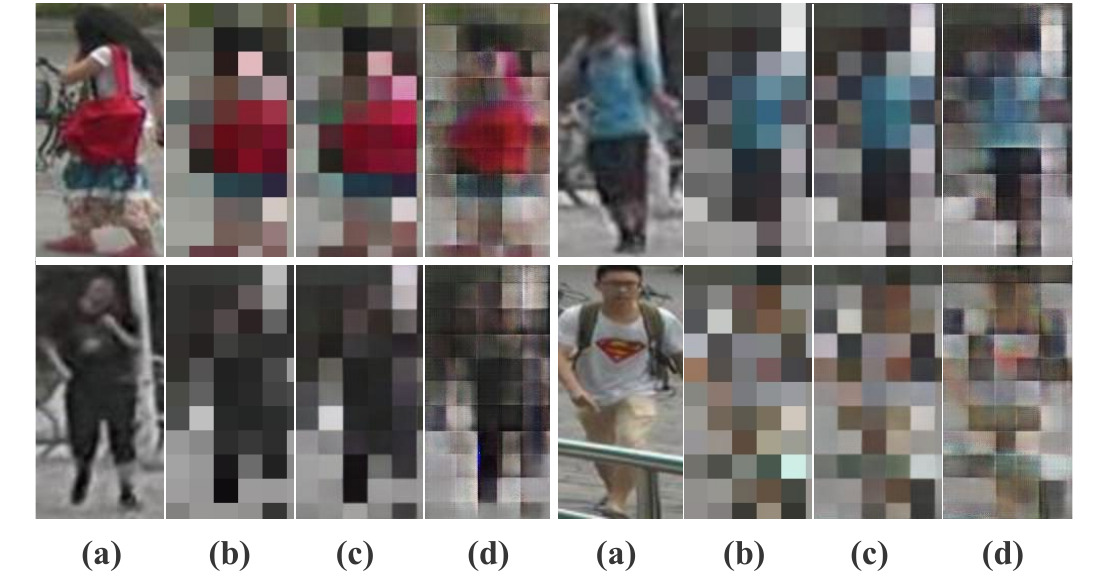}
  
  \caption{{Qualitative comparison on supervision upgradation. (a) raw images; (b)pixelated images; (c) anonymized images w/o U; (d) anonymized images w/ U.}}
  
\label{fig:upgrade-results}
\end{figure}

\renewcommand\arraystretch{0.87}
\begin{table*}[t]
  \centering
 \caption{\label{tab:reid}Evaluation of Re-ID performance on three Re-ID datasets. ``Base'' means AGW model trained on desensitized images. ``Upper'' indicates original AGW model. Rank at $r$ accuracy(\%), mAP (\%) and mINP (\%) are reported.}
\begin{tabular}{c|cc|cc|cccc|cccc|cccc}\hline
& \multicolumn{2}{c|}{query} 
& \multicolumn{2}{c|}{gallery} &\multicolumn{4}{c|}{Market-1501}  & \multicolumn{4}{c|}{MSMT17}  & \multicolumn{4}{c}{CUHK03}\\\hline
 & OI & PI & OI & PI & $r=1$   & $r=5 $ &  mAP & mINP     & $r=1$ & $r=5$ &  mAP & mINP  & $r=1$ & $r=5$ &  mAP & mINP  \\\hline

\multicolumn{7}{l}{\textit{(a) Evaluation of blurring.}}\\\hline
\multirow{4}{*}{\shortstack{ Base}}
& $\checkmark$ & & $\checkmark$&  & 84.8 & 94.1 & 67.4 & 32.2 
  & 30.5 & 44.0 & 17.1  & 2.8 & 30.4 & 50.4 & 31.5 & 22.3 \\
& $\checkmark$ & & & $\checkmark$ & 40.1 & 59.4 & 25.4 & 6.3 
  & 21.3 & 35.1 & 10.7 & 1.4 &	14.6 &	28.4 & 14.8 & 8.6 \\
& & $\checkmark$ & $\checkmark$ & & 18.3 & 31.7 & 15.5 & 5.2 
  &	16.2 &	28.0 & 9.4 & 1.5 & 10.4 & 20.4 & 12.4 &	8.4  \\
& & $\checkmark$ & &  $\checkmark$  & 67.3 & 83.5 & 44.2 & 13.7 
  & 15.2 & 24.3 & 7.2 & 0.8 & 8.2 & 19.8 &	10.7 &	6.9\\\hline
  
 \multirow{4}{*}{\shortstack{Ours\\(w/o U)}}
& $\checkmark$ & & $\checkmark$&  & 83.1  & 93.7 & 61.9 & 24.8 
  & 43.6 & 57.6 & 24.0 & 3.7 &	31.6& 52.6 & 32.5 & 23.0 \\
& $\checkmark$ & & & $\checkmark$ & 68.3 & 84.9 & 45.9  & 14.6
  & 28.4 & 42.3 & 9.1 & 1.2 & 13.8 & 25.8 & 14.8 & 9.0 \\
& & $\checkmark$ & $\checkmark$ & & 46.8 & 67.7 & 34.2 & 11.7
  & 10.0 & 19.2 & 5.8 &	0.8 &	8.9 & 19.4 & 10.8 & 7.1 \\
& & $\checkmark$ & &  $\checkmark$  & 75.8 & 88.9 & 52.4 & 18.3 
  &	14.7 &	23.6 & 6.0 & 0.5 & 14.3 & 28.5 & 15.0 & 8.9 \\\hline

\multirow{4}{*}{{ Ours}}
& $\checkmark$ & & $\checkmark$&  & \textbf{91.6}  & \textbf{97.4} & \textbf{79.4} & \textbf{47.4} 
  & \textbf{51.5} & \textbf{65.3} & \textbf{31.1} & \textbf{6.0} &	\textbf{41.9} &	\textbf{62.0} &	\textbf{41.7} & \textbf{30.4} \\
& $\checkmark$ & & & $\checkmark$ & 88.2 & 95.8 & 72.0 & 37.0
  & 51.1 & 64.9 & 29.7 & 5.2 &	39.2 &	59.3 & 38.4 & 27.2 \\
& & $\checkmark$ & $\checkmark$ & & 82.5 & 93.6 & 67.5 & 36.0
  & 50.5 & 64.7 & 30.5 & 5.7 &	35.3 &	55.4 &	35.5 & 25.4 \\
& & $\checkmark$ & &  $\checkmark$  & 89.2 & 96.4 & 74.3 & 39.4 
  &	48.7 & 62.4 & 28.5 & 4.9 & 33.2 & 55.3 & 34.7 &	25.0\\\hline
  
\multicolumn{7}{l}{\textit{(b) Evaluation of pixelation.}}\\\hline
\multirow{4}{*}{{ Base}}
& $\checkmark$ & & $\checkmark$&  & 87.4  & 96.1 & 73.4 & 39.5 
  & 25.0 & 38.3 & 15.3 & 2.6 & 28.5 & 50.0 & 31.5 & 22.9\\
& $\checkmark$ & & & $\checkmark$ & 75.3 & 91.1 & 53.6 & 17.2
  & 16.3 & 29.0 & 8.7 &	1.0 & 17.7 & 36.5 & 17.6 & 9.1 \\
& & $\checkmark$ & $\checkmark$ & & 70.9 & 86.4 & 54.7 & 24.1
  & 14.6 & 24.7 & 9.0 &	1.6 & 15.1 & 29.5 & 17.7 & 12.3 \\
& & $\checkmark$ & &  $\checkmark$  & 64.3 & 83.5 & 43.4 & 13.0 
  &	10.6 &	19.7 & 5.7 & 0.7 & 8.8 & 20.3 &	9.9 & 5.3\\\hline

 \multirow{4}{*}{\shortstack{Ours\\(w/o U)}}
& $\checkmark$ & & $\checkmark$&  & 86.3  & 95.3 & 69.8 & 34.0 
  & 34.3 & 47.7 & 18.8 & 2.8 &	24.9 &	44.9 & 27.3 & 19.1 \\
& $\checkmark$ & & & $\checkmark$ & 80.1 & 92.6 & 57.4  & 18.6
  & 26.2 & 40.4 & 12.3 & 1.3 &	24.2 &	44.8 & 23.3 & 13.6 \\
& & $\checkmark$ & $\checkmark$ & & 75.1 & 89.1 & 57.1 & 24.3
  & 24.6 & 37.0 &	12.9 &	1.9 & 19.6 & 35.1 & 20.8 & 14.1 \\
& & $\checkmark $ & &  $\checkmark$  & 73.2 & 89.1 & 49.7 & 15.7 
  &	20.5 &	32.4 & 9.3 & 1.0 & 12.1 & 25.9 & 13.2 &	7.5\\\hline
  
\multirow{4}{*}{{ Ours}}
& $\checkmark$ & & $\checkmark$&  & \textbf{89.4}  & \textbf{96.2} & \textbf{75.4} & \textbf{42.4} 
  & 48.6 & 63.2 & \textbf{29.8} & \textbf{5.9} &	\textbf{38.8} &	\textbf{64.3} & \textbf{42.4} & 31.6 \\
& $\checkmark$ & & & $\checkmark$ & 88.5 & 95.7 & 71.9  & 35.9
  & \textbf{49.1} & \textbf{63.6} & 29.3 & 5.5 &	37.8 &	60.5 & 41.4 & \textbf{32.2} \\
& & $\checkmark$ & $\checkmark$ & & 86.8 & 94.9 & 72.3 & 39.1
  & 48.5 & 63.6 & 29.8 & 5.7 &	30.4 &	50.6 & 30.6 & 20.7 \\
& & $\checkmark$ & &  $\checkmark$  & 87.0 & 95.6 & 70.5 & 34.8 
  &	48.1 &	62.7 & 29.3 & 5.6 & 27.6 & 47.3 & 28.5 & 19.2 \\\hline
  
\multicolumn{7}{l}{\textit{(c) Evaluation of Gaussian noise.}}\\\hline
\multirow{4}{*}{{ Base}}
& $\checkmark$ & & $\checkmark$&  & 75.9  & 89.4 & 51.5 & 16.5 
  & 24.0 & 34.2 & 11.1 & 1.3 & 14.0 & 27.4 & 15.7 & 9.9\\
& $\checkmark$ & & & $\checkmark$ & 50.8 & 70.3 & 30.2 & 5.8
  & 20.4 & 32.4 & 8.6 &	0.9 & 9.1 & 19.4 & 9.9 & 5.4 \\
& & $\checkmark$ & $\checkmark$ & & 41.7 & 62.9 & 26.5 & 6.8
  & 18.5 & 28.4 & 8.4 &	0.9 & 8.6 & 18.9 & 9.9 & 5.9 \\
& & $\checkmark$ & &  $\checkmark$  & 68.7 & 85.9 & 43.2 & 11.9 
  &	18.2 &	27.6 & 7.8 & 0.8 & 8.1 & 18.7 &	10.2 & 5.9\\\hline

 \multirow{4}{*}{\shortstack{Ours\\(w/o U)}}
& $\checkmark$ & & $\checkmark$&  & 90.4  & 96.7 & 75.9 & 41.9 
  & 41.9 & 55.6 & 24.0 & 4.2 &	28.1 &	47.6 & 30.2 & 21.8 \\
& $\checkmark$ & & & $\checkmark$ & 77.5 & 89.7 & 57.3  & 21.0
  & 36.0 & 51.5 & 18.4 & 2.7 &	30.4 &	51.1 & 30.2 & 20.0 \\
& & $\checkmark$ & $\checkmark$ & & 67.5 & 82.0 & 50.8 & 20.7
  & 29.4 & 41.4 & 15.7 & 2.5 & 30.4 & 50.4 & 29.9 & 20.3 \\
& & $\checkmark $ & &  $\checkmark$  & 84.4 & 93.6 & 62.8 & 25.8 
  & 31.3 & 43.4 & 15.0 & 1.9 & 31.9 & 53.2 & 32.4 &	22.8\\\hline
  
\multirow{4}{*}{{ Ours}}
& $\checkmark$ & & $\checkmark$&  & \textbf{91.7} & 96.8 & \textbf{78.2} & \textbf{45.4} &46.9 & 60.7 & \textbf{27.6} & \textbf{4.9} &35.8 & 56.1 & 36.8 & 26.7  \\
& $\checkmark$ & & & $\checkmark$ & 83.5 & 92.8 & 68.0 & 33.6 &\textbf{48.1} & \textbf{62.7} & 27.3 & 4.4 &36.4 & 57.3 & 36.3 & 25.6  \\
& & $\checkmark$ & $\checkmark$ & & 83.8 & 92.7 & 67.3 & 32.4 &46.2 & 59.4 & 26.1 & 4.4 &37.9 & 57.1 & 36.9 & 25.5  \\
& & $\checkmark$ & &  $\checkmark$  & 91.2 & \textbf{96.9} & 77.0 & 44.3 &46.0 & 59.4 & 26.0 & 4.4 &\textbf{41.9} & \textbf{62.4} & \textbf{41.6} & \textbf{30.4}\\\hline
 Upper &$\checkmark$  & & $\checkmark$ & & 95.7 & 98.4 & 88.6 &	66.7  &	68.6 & 79.7 & 49.8 & 15.0 & 67.3 & 82.8 & 65.8 & 54.6 \\\hline
 \end{tabular}
\end{table*}

\textbf{Effect of Supervision Upgradation.} As shown in Table~\ref{tab:reid}, compared to our model without upgradation, the model with upgradation generally performs better under all metrics and settings, e.g., the average Rank-1 increases 19.4\% under the evaluation of blurring on Market1501 dataset. This shows that utilizing supervision upgradation indeed helps in improving Re-ID performance on our anonymized images. Fig.~\ref{fig:upgrade-results} shows that the anonymized images with upgradation retain a strong visual obfuscation effect. Instead of those without upgradation which follow the style of pixelation, raw images are obfuscated in a different learned style.

\textbf{Discussion.}
In our privacy-preserving system, given a frame of raw video, pedestrians can be anonymized based on the detected bounding boxes. These anonymized bounding boxes can further be used to recover original raw images by police officers and adopted as public datasets for researchers. However, given an anonymized frame, anonymized pedestrians are unable to be detected by standard person detectors without training on them. It needs further joint learning with pedestrian detection tasks.

%-------------------------------------------------------------------------

\section{Conclusion}
This paper proposes a new reversible anonymization framework to explore the privacy-utility trade-off for pedestrian images from Re-ID perspective, which can reversibly generate full-body anonymous images with little performance degradation in Re-ID tasks. We further propose a progressive training strategy to improve the Re-ID performance. Extensive experiments further demonstrate the effectiveness of our method using anonymized pedestrian images for privacy protection, recovery, and person re-identification.

\section*{ACKNOWLEDGMENTS}
This work is supported by National Natural Science Foundation of China (62176188), Key Research and Development Program of Hubei Province (2021BAA187), Special Fund of Hubei Luojia Laboratory (220100015), Zhejiang Lab (NO.2022NF0AB01).

\section{Appendix}
\begin{figure*}[t]
\centering
  \includegraphics[ width=17cm]{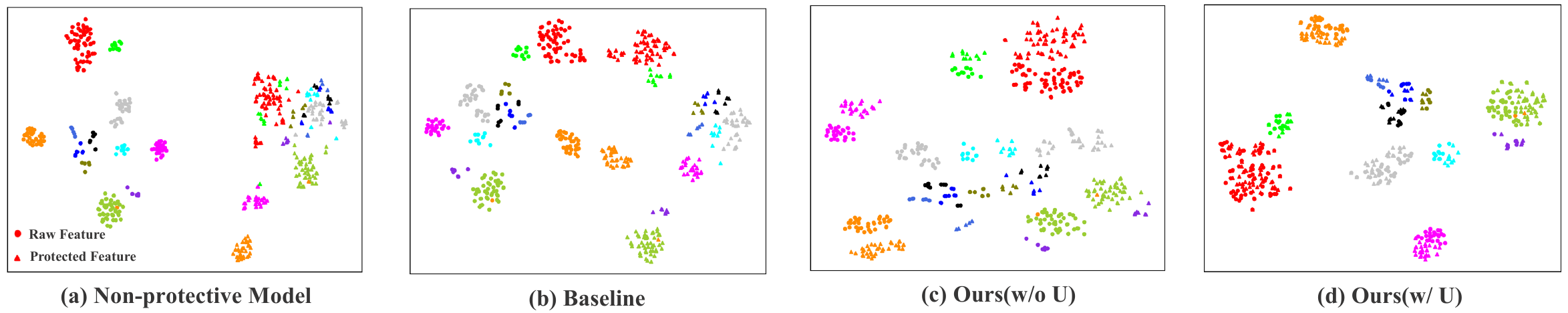}
  \caption{\small{Comparison of the feature distribution extracted by (a) non-protective Re-ID model, (b) the baseline blurring-based model, (c) ours (w/o U), and (d) ours (w/ U) on Market1501 dataset. Features with the same color are from images of the same person.}}
\label{fig:tsne}
\end{figure*}

\subsection*{A. Feature Distribution Comparison}
 In order to show more intuitively that our anonymized images are suitable as the public dataset for Re-ID research, we visualize the feature distribution of both raw and protected images in Fig.~\ref{fig:tsne}. The features are produced from the same batch of test samples and are extracted from the current non-protective Re-ID model, the baseline blurring-based model, and our blurring-based Re-ID model with/without supervision upgradation. If the protected feature distribution can be clustered distinctly by classes, and is similar to the original raw feature distribution, then the protected images should be suitable as a public Re-ID dataset. 

 \textit{a) Non-protective Model}: Fig.~\ref{fig:tsne}(a) illustrates the feature distribution extracted from the current non-protective Re-ID model AGW \cite{ye2021deep} which is trained on only raw images and tested on both raw and blurred images. It can be clearly seen that raw features are clustered by classes while protected features of different classes are mostly mixed together, indicating the Re-ID model trained on only raw images achieves poor performance when testing images contain protected images. 
 
 \textit{b) Baseline}: The baseline Re-ID model is trained and tested on paired raw and blurred images. As illustrated in Fig.~\ref{fig:tsne}(b), compared to non-protective model, the protected features are clustered better and the distance between raw and protected feature distribution is narrowed. This indicates that the Re-ID performance is improved when the test set consists of protected images. However, the blurred images are not able to replace raw images as a Re-ID dataset due to the large deviation of feature distribution after blurring. 
 
 \textit{c) Ours (w/o U)}: Fig.~\ref{fig:tsne}(c) shows the feature distribution extracted from our model that is jointly trained without supervision upgradation and tested on raw and anonymized images. Compared to the baseline, the raw and protected feature distributions are significantly pulled in. However, there still exists observable misalignment between these two feature distributions. 
 
 \textit{d) Ours (w/ U)}: To further narrow the distance between these two feature distributions, we propose a training strategy, i.e., progressive supervision upgradation. As shown in Fig.~\ref{fig:tsne}(d), compared to baseline and our method w/o U, our method with supervision upgradation eliminates the deviation between these two types of feature distribution while retaining the distinction between classes. Compared to raw feature distribution from non-protective model, our method achieves comparable performance of forming clusters. The results indicate that the model trained on raw and our anonymized images can perform well under original, protected, and crossed settings and our anonymized images are suitable as the public Re-ID dataset.

\subsection*{B. Results of Privacy Protection and Recovery}
\begin{figure*}[t]
\centering
  \includegraphics[ width=13cm]{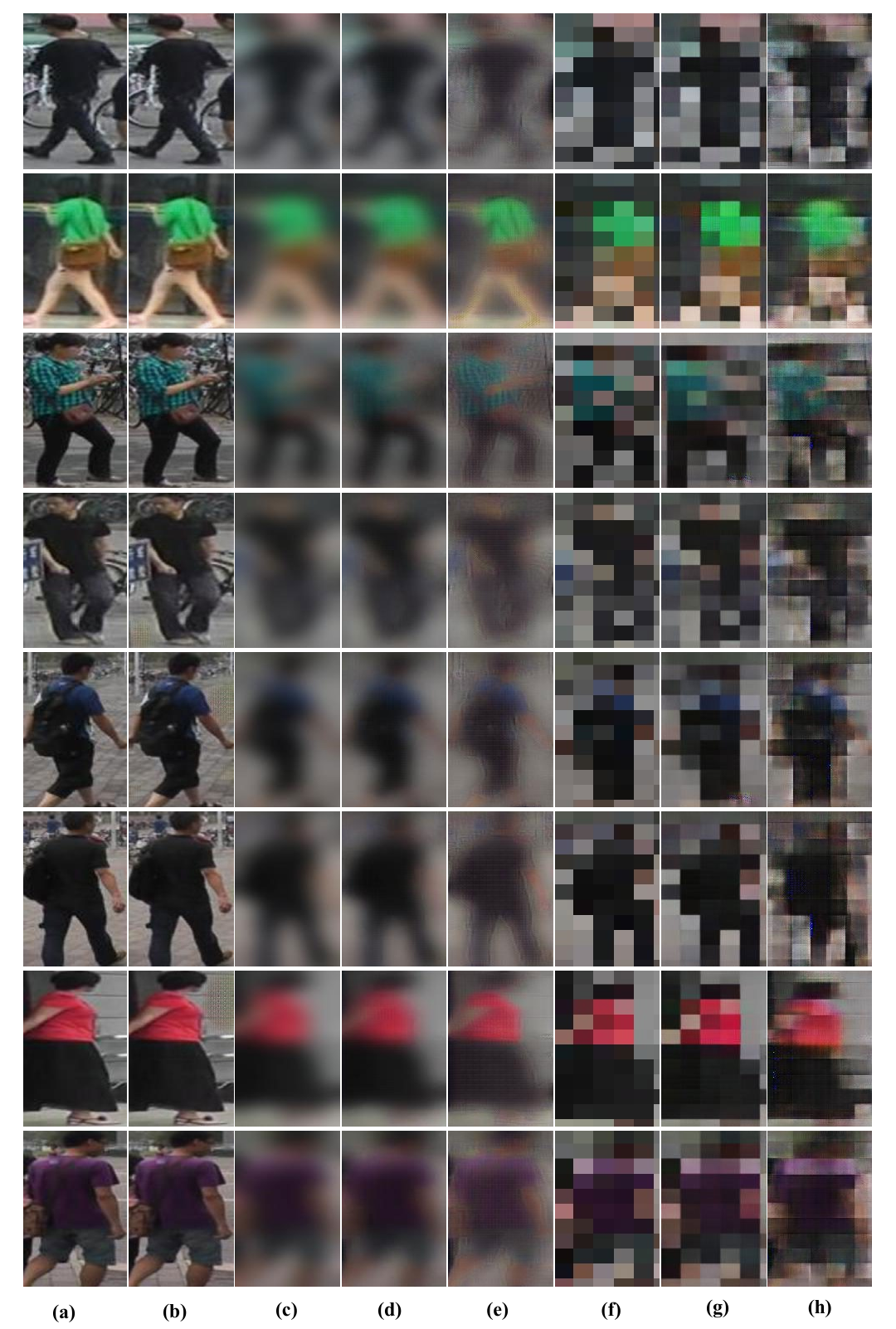}
  \caption{\small{Qualitative comparison between baseline and our method. (a)/(b) denote raw/recovered images; (c) indicates blurred images; (d)/(e) represent anonymized images guided by blurred images without/with supervision upgradation; (f)/(g)/(h) are similar to (c)/(d)/(e) with blurred images being replaced by pixelated images. }}
\label{fig:results}
\end{figure*}

In Fig.~\ref{fig:qualitative} and Fig.~\ref{fig:recovery} of the main paper, we separately showed qualitative results of privacy protection and recovery. Besides, in Fig.~\ref{fig:upgrade-results} of the main paper, we performed comparison on supervision upgradation. In this part, we combine the three qualitative experiments and show more images in Fig.~\ref{fig:results}. It can be seen that our anonymized images achieve good visual obfuscation effect and our recovered images are visually similar to raw images.

% \begin{acks}
% \end{acks}

\bibliographystyle{ACM-Reference-Format}
\bibliography{reference}


\begin{thebibliography}{44}

%%% the Web address.

\ifx \showCODEN    \undefined \def \showCODEN     #1{\unskip}     \fi
\ifx \showDOI      \undefined \def \showDOI       #1{#1}\fi
\ifx \showISBNx    \undefined \def \showISBNx     #1{\unskip}     \fi
\ifx \showISBNxiii \undefined \def \showISBNxiii  #1{\unskip}     \fi
\ifx \showISSN     \undefined \def \showISSN      #1{\unskip}     \fi
\ifx \showLCCN     \undefined \def \showLCCN      #1{\unskip}     \fi
\ifx \shownote     \undefined \def \shownote      #1{#1}          \fi
\ifx \showarticletitle \undefined \def \showarticletitle #1{#1}   \fi
\ifx \showURL      \undefined \def \showURL       {\relax}        \fi
% invisible to TeX
\providecommand\bibfield[2]{#2}
\providecommand\bibinfo[2]{#2}
\providecommand\natexlab[1]{#1}
\providecommand\showeprint[2][]{arXiv:#2}

\bibitem[Chen et~al\mbox{.}(2021)]%
        {chen2021perceptual}
\bibfield{author}{\bibinfo{person}{Jia-Wei Chen}, \bibinfo{person}{Li-Ju Chen},
  \bibinfo{person}{Chia-Mu Yu}, {and} \bibinfo{person}{Chun-Shien Lu}.}
  \bibinfo{year}{2021}\natexlab{}.
\newblock \showarticletitle{Perceptual Indistinguishability-Net (PI-Net):
  Facial Image Obfuscation with Manipulable Semantics}. In
  \bibinfo{booktitle}{\emph{CVPR}}. \bibinfo{pages}{6478--6487}.
\newblock


\bibitem[Deepfake(2020)]%
        {deepfake}
\bibfield{author}{\bibinfo{person}{Deepfake}.} \bibinfo{year}{2020}\natexlab{}.
\newblock \bibinfo{title}{Deepfakes faceswap}.
\newblock
\newblock
\newblock
\shownote{\url{https://github.com/deepfakes/faceswap}}.


\bibitem[Deng et~al\mbox{.}(2009)]%
        {deng2009imagenet}
\bibfield{author}{\bibinfo{person}{Jia Deng}, \bibinfo{person}{Wei Dong},
  \bibinfo{person}{Richard Socher}, \bibinfo{person}{Li-Jia Li},
  \bibinfo{person}{Kai Li}, {and} \bibinfo{person}{Li Fei-Fei}.}
  \bibinfo{year}{2009}\natexlab{}.
\newblock \showarticletitle{Imagenet: A large-scale hierarchical image
  database}. In \bibinfo{booktitle}{\emph{CVPR}}. \bibinfo{pages}{248--255}.
\newblock


\bibitem[Dietlmeier et~al\mbox{.}(2022)]%
        {dietlmeier2022improving}
\bibfield{author}{\bibinfo{person}{Julia Dietlmeier}, \bibinfo{person}{Feiyan
  Hu}, \bibinfo{person}{Frances Ryan}, \bibinfo{person}{Noel~E O'Connor}, {and}
  \bibinfo{person}{Kevin McGuinness}.} \bibinfo{year}{2022}\natexlab{}.
\newblock \showarticletitle{Improving Person Re-Identification with Temporal
  Constraints}. In \bibinfo{booktitle}{\emph{CVPR}}. \bibinfo{pages}{540--549}.
\newblock


\bibitem[Dwork(2008)]%
        {dwork2008differential}
\bibfield{author}{\bibinfo{person}{Cynthia Dwork}.}
  \bibinfo{year}{2008}\natexlab{}.
\newblock \showarticletitle{Differential privacy: A survey of results}. In
  \bibinfo{booktitle}{\emph{TAMC}}. \bibinfo{pages}{1--19}.
\newblock


\bibitem[Dwork et~al\mbox{.}(2006)]%
        {dwork2006calibrating}
\bibfield{author}{\bibinfo{person}{Cynthia Dwork}, \bibinfo{person}{Frank
  McSherry}, \bibinfo{person}{Kobbi Nissim}, {and} \bibinfo{person}{Adam
  Smith}.} \bibinfo{year}{2006}\natexlab{}.
\newblock \showarticletitle{Calibrating noise to sensitivity in private data
  analysis}. In \bibinfo{booktitle}{\emph{TCC}}. \bibinfo{pages}{265--284}.
\newblock


\bibitem[Facebook(2021)]%
        {facebook}
\bibfield{author}{\bibinfo{person}{Facebook}.} \bibinfo{year}{2021}\natexlab{}.
\newblock \bibinfo{title}{An Update On Our Use of Face Recognition}.
\newblock
\newblock
\newblock
\shownote{\url{https://about.fb.com/news/2021/11/update-on-use-of-face-recognition}}.


\bibitem[Gafni et~al\mbox{.}(2019)]%
        {gafni2019live}
\bibfield{author}{\bibinfo{person}{Oran Gafni}, \bibinfo{person}{Lior Wolf},
  {and} \bibinfo{person}{Yaniv Taigman}.} \bibinfo{year}{2019}\natexlab{}.
\newblock \showarticletitle{Live face de-identification in video}. In
  \bibinfo{booktitle}{\emph{ICCV}}. \bibinfo{pages}{9378--9387}.
\newblock


\bibitem[Goodfellow et~al\mbox{.}(2014)]%
        {Goodfellow2014GenerativeAN}
\bibfield{author}{\bibinfo{person}{Ian~J. Goodfellow}, \bibinfo{person}{Jean
  Pouget-Abadie}, \bibinfo{person}{Mehdi Mirza}, \bibinfo{person}{Bing Xu},
  \bibinfo{person}{David Warde-Farley}, \bibinfo{person}{Sherjil Ozair},
  \bibinfo{person}{Aaron~C. Courville}, {and} \bibinfo{person}{Yoshua Bengio}.}
  \bibinfo{year}{2014}\natexlab{}.
\newblock \showarticletitle{Generative Adversarial Nets}. In
  \bibinfo{booktitle}{\emph{NIPS}}.
\newblock


\bibitem[He et~al\mbox{.}(2016)]%
        {he2016deep}
\bibfield{author}{\bibinfo{person}{Kaiming He}, \bibinfo{person}{Xiangyu
  Zhang}, \bibinfo{person}{Shaoqing Ren}, {and} \bibinfo{person}{Jian Sun}.}
  \bibinfo{year}{2016}\natexlab{}.
\newblock \showarticletitle{Deep residual learning for image recognition}. In
  \bibinfo{booktitle}{\emph{CVPR}}. \bibinfo{pages}{770--778}.
\newblock


\bibitem[Hukkel{\aa}s et~al\mbox{.}(2019)]%
        {hukkelaas2019deepprivacy}
\bibfield{author}{\bibinfo{person}{H{\aa}kon Hukkel{\aa}s},
  \bibinfo{person}{Rudolf Mester}, {and} \bibinfo{person}{Frank Lindseth}.}
  \bibinfo{year}{2019}\natexlab{}.
\newblock \showarticletitle{Deepprivacy: A generative adversarial network for
  face anonymization}. In \bibinfo{booktitle}{\emph{ISVC}}.
  \bibinfo{pages}{565--578}.
\newblock


\bibitem[Isola et~al\mbox{.}(2017)]%
        {isola2017image}
\bibfield{author}{\bibinfo{person}{Phillip Isola}, \bibinfo{person}{Jun-Yan
  Zhu}, \bibinfo{person}{Tinghui Zhou}, {and} \bibinfo{person}{Alexei~A
  Efros}.} \bibinfo{year}{2017}\natexlab{}.
\newblock \showarticletitle{Image-to-image translation with conditional
  adversarial networks}. In \bibinfo{booktitle}{\emph{CVPR}}.
  \bibinfo{pages}{1125--1134}.
\newblock


\bibitem[Kairouz et~al\mbox{.}(2019)]%
        {kairouz2019advances}
\bibfield{author}{\bibinfo{person}{Peter Kairouz}, \bibinfo{person}{H~Brendan
  McMahan}, \bibinfo{person}{Brendan Avent}, \bibinfo{person}{Aur{\'e}lien
  Bellet}, \bibinfo{person}{Mehdi Bennis}, \bibinfo{person}{Arjun~Nitin
  Bhagoji}, \bibinfo{person}{Kallista Bonawitz}, \bibinfo{person}{Zachary
  Charles}, \bibinfo{person}{Graham Cormode}, \bibinfo{person}{Rachel
  Cummings}, {et~al\mbox{.}}} \bibinfo{year}{2019}\natexlab{}.
\newblock \showarticletitle{Advances and open problems in federated learning}.
\newblock \bibinfo{journal}{\emph{arXiv preprint arXiv:1912.04977}}
  (\bibinfo{year}{2019}).
\newblock


\bibitem[Kingma and Ba(2014)]%
        {kingma2014adam}
\bibfield{author}{\bibinfo{person}{Diederik~P Kingma} {and}
  \bibinfo{person}{Jimmy Ba}.} \bibinfo{year}{2014}\natexlab{}.
\newblock \showarticletitle{Adam: A method for stochastic optimization}.
\newblock \bibinfo{journal}{\emph{arXiv preprint arXiv:1412.6980}}
  (\bibinfo{year}{2014}).
\newblock


\bibitem[Kuang et~al\mbox{.}(2021a)]%
        {kuang2021unnoticeable}
\bibfield{author}{\bibinfo{person}{Zhenzhong Kuang}, \bibinfo{person}{Zhiqiang
  Guo}, \bibinfo{person}{Jinglong Fang}, \bibinfo{person}{Jun Yu},
  \bibinfo{person}{Noboru Babaguchi}, {and} \bibinfo{person}{Jianping Fan}.}
  \bibinfo{year}{2021}\natexlab{a}.
\newblock \showarticletitle{Unnoticeable synthetic face replacement for image
  privacy protection}.
\newblock \bibinfo{journal}{\emph{Neurocomputing}}  \bibinfo{volume}{457}
  (\bibinfo{year}{2021}), \bibinfo{pages}{322--333}.
\newblock


\bibitem[Kuang et~al\mbox{.}(2021b)]%
        {kuang2021effective}
\bibfield{author}{\bibinfo{person}{Zhenzhong Kuang}, \bibinfo{person}{Huigui
  Liu}, \bibinfo{person}{Jun Yu}, \bibinfo{person}{Aikui Tian},
  \bibinfo{person}{Lei Wang}, \bibinfo{person}{Jianping Fan}, {and}
  \bibinfo{person}{Noboru Babaguchi}.} \bibinfo{year}{2021}\natexlab{b}.
\newblock \showarticletitle{Effective De-identification Generative Adversarial
  Network for Face Anonymization}. In \bibinfo{booktitle}{\emph{ACM MM}}.
  \bibinfo{pages}{3182--3191}.
\newblock


\bibitem[Li and Lin(2019)]%
        {li2019anonymousnet}
\bibfield{author}{\bibinfo{person}{Tao Li} {and} \bibinfo{person}{Lei Lin}.}
  \bibinfo{year}{2019}\natexlab{}.
\newblock \showarticletitle{Anonymousnet: Natural face de-identification with
  measurable privacy}. In \bibinfo{booktitle}{\emph{CVPR}}.
  \bibinfo{pages}{0--0}.
\newblock


\bibitem[Luo et~al\mbox{.}(2019)]%
        {luo2019strong}
\bibfield{author}{\bibinfo{person}{Hao Luo}, \bibinfo{person}{Wei Jiang},
  \bibinfo{person}{Youzhi Gu}, \bibinfo{person}{Fuxu Liu},
  \bibinfo{person}{Xingyu Liao}, \bibinfo{person}{Shenqi Lai}, {and}
  \bibinfo{person}{Jianyang Gu}.} \bibinfo{year}{2019}\natexlab{}.
\newblock \showarticletitle{A strong baseline and batch normalization neck for
  deep person re-identification}.
\newblock \bibinfo{journal}{\emph{ACM MM}} \bibinfo{volume}{22},
  \bibinfo{number}{10} (\bibinfo{year}{2019}), \bibinfo{pages}{2597--2609}.
\newblock


\bibitem[Maximov et~al\mbox{.}(2020)]%
        {maximov2020ciagan}
\bibfield{author}{\bibinfo{person}{Maxim Maximov}, \bibinfo{person}{Ismail
  Elezi}, {and} \bibinfo{person}{Laura Leal-Taix{\'e}}.}
  \bibinfo{year}{2020}\natexlab{}.
\newblock \showarticletitle{Ciagan: Conditional identity anonymization
  generative adversarial networks}. In \bibinfo{booktitle}{\emph{CVPR}}.
  \bibinfo{pages}{5447--5456}.
\newblock


\bibitem[McMahan et~al\mbox{.}(2017)]%
        {mcmahan2017communication}
\bibfield{author}{\bibinfo{person}{Brendan McMahan}, \bibinfo{person}{Eider
  Moore}, \bibinfo{person}{Daniel Ramage}, \bibinfo{person}{Seth Hampson},
  {and} \bibinfo{person}{Blaise~Aguera y Arcas}.}
  \bibinfo{year}{2017}\natexlab{}.
\newblock \showarticletitle{Communication-efficient learning of deep networks
  from decentralized data}. In \bibinfo{booktitle}{\emph{AISTATS}}.
  \bibinfo{pages}{1273--1282}.
\newblock


\bibitem[Proen{\c{c}}a(2020)]%
        {proencca2020uu}
\bibfield{author}{\bibinfo{person}{Hugo Proen{\c{c}}a}.}
  \bibinfo{year}{2020}\natexlab{}.
\newblock \showarticletitle{The uu-net: Reversible face de-identification for
  visual surveillance video footage}.
\newblock \bibinfo{journal}{\emph{arXiv preprint arXiv:2007.04316}}
  (\bibinfo{year}{2020}).
\newblock


\bibitem[Ren et~al\mbox{.}(2018)]%
        {ren2018learning}
\bibfield{author}{\bibinfo{person}{Zhongzheng Ren}, \bibinfo{person}{Yong~Jae
  Lee}, {and} \bibinfo{person}{Michael~S Ryoo}.}
  \bibinfo{year}{2018}\natexlab{}.
\newblock \showarticletitle{Learning to anonymize faces for privacy preserving
  action detection}. In \bibinfo{booktitle}{\emph{ECCV}}.
  \bibinfo{pages}{620--636}.
\newblock


\bibitem[Ristani et~al\mbox{.}(2016)]%
        {ristani2016performance}
\bibfield{author}{\bibinfo{person}{Ergys Ristani}, \bibinfo{person}{Francesco
  Solera}, \bibinfo{person}{Roger Zou}, \bibinfo{person}{Rita Cucchiara}, {and}
  \bibinfo{person}{Carlo Tomasi}.} \bibinfo{year}{2016}\natexlab{}.
\newblock \showarticletitle{Performance measures and a data set for
  multi-target, multi-camera tracking}. In \bibinfo{booktitle}{\emph{ECCV}}.
  \bibinfo{pages}{17--35}.
\newblock


\bibitem[Sun et~al\mbox{.}(2018a)]%
        {sun2018natural}
\bibfield{author}{\bibinfo{person}{Qianru Sun}, \bibinfo{person}{Liqian Ma},
  \bibinfo{person}{Seong~Joon Oh}, \bibinfo{person}{Luc Van~Gool},
  \bibinfo{person}{Bernt Schiele}, {and} \bibinfo{person}{Mario Fritz}.}
  \bibinfo{year}{2018}\natexlab{a}.
\newblock \showarticletitle{Natural and effective obfuscation by head
  inpainting}. In \bibinfo{booktitle}{\emph{CVPR}}.
  \bibinfo{pages}{5050--5059}.
\newblock


\bibitem[Sun et~al\mbox{.}(2018b)]%
        {sun2018hybrid}
\bibfield{author}{\bibinfo{person}{Qianru Sun}, \bibinfo{person}{Ayush Tewari},
  \bibinfo{person}{Weipeng Xu}, \bibinfo{person}{Mario Fritz},
  \bibinfo{person}{Christian Theobalt}, {and} \bibinfo{person}{Bernt Schiele}.}
  \bibinfo{year}{2018}\natexlab{b}.
\newblock \showarticletitle{A hybrid model for identity obfuscation by face
  replacement}. In \bibinfo{booktitle}{\emph{ECCV}}. \bibinfo{pages}{553--569}.
\newblock


\bibitem[Torralba et~al\mbox{.}(2008)]%
        {torralba200880}
\bibfield{author}{\bibinfo{person}{Antonio Torralba}, \bibinfo{person}{Rob
  Fergus}, {and} \bibinfo{person}{William~T Freeman}.}
  \bibinfo{year}{2008}\natexlab{}.
\newblock \showarticletitle{80 million tiny images: A large data set for
  nonparametric object and scene recognition}.
\newblock \bibinfo{journal}{\emph{IEEE TPAMI}} \bibinfo{volume}{30},
  \bibinfo{number}{11} (\bibinfo{year}{2008}), \bibinfo{pages}{1958--1970}.
\newblock


\bibitem[Wang et~al\mbox{.}(2018)]%
        {wang2018learning}
\bibfield{author}{\bibinfo{person}{Guanshuo Wang}, \bibinfo{person}{Yufeng
  Yuan}, \bibinfo{person}{Xiong Chen}, \bibinfo{person}{Jiwei Li}, {and}
  \bibinfo{person}{Xi Zhou}.} \bibinfo{year}{2018}\natexlab{}.
\newblock \showarticletitle{Learning discriminative features with multiple
  granularities for person re-identification}. In \bibinfo{booktitle}{\emph{ACM
  MM}}. \bibinfo{pages}{274--282}.
\newblock


\bibitem[Wang et~al\mbox{.}(2007)]%
        {wang2007shape}
\bibfield{author}{\bibinfo{person}{Xiaogang Wang}, \bibinfo{person}{Gianfranco
  Doretto}, \bibinfo{person}{Thomas Sebastian}, \bibinfo{person}{Jens
  Rittscher}, {and} \bibinfo{person}{Peter Tu}.}
  \bibinfo{year}{2007}\natexlab{}.
\newblock \showarticletitle{Shape and appearance context modeling}. In
  \bibinfo{booktitle}{\emph{ICCV}}. \bibinfo{pages}{1--8}.
\newblock


\bibitem[Wang et~al\mbox{.}(2004)]%
        {wang2004image}
\bibfield{author}{\bibinfo{person}{Zhou Wang}, \bibinfo{person}{Alan~C Bovik},
  \bibinfo{person}{Hamid~R Sheikh}, {and} \bibinfo{person}{Eero~P Simoncelli}.}
  \bibinfo{year}{2004}\natexlab{}.
\newblock \showarticletitle{Image quality assessment: from error visibility to
  structural similarity}.
\newblock \bibinfo{journal}{\emph{IEEE TIP}} \bibinfo{volume}{13},
  \bibinfo{number}{4} (\bibinfo{year}{2004}), \bibinfo{pages}{600--612}.
\newblock


\bibitem[Wei et~al\mbox{.}(2018)]%
        {wei2018person}
\bibfield{author}{\bibinfo{person}{Longhui Wei}, \bibinfo{person}{Shiliang
  Zhang}, \bibinfo{person}{Wen Gao}, {and} \bibinfo{person}{Qi Tian}.}
  \bibinfo{year}{2018}\natexlab{}.
\newblock \showarticletitle{Person transfer gan to bridge domain gap for person
  re-identification}. In \bibinfo{booktitle}{\emph{CVPR}}.
  \bibinfo{pages}{79--88}.
\newblock


\bibitem[Wen et~al\mbox{.}(2016)]%
        {wen2016discriminative}
\bibfield{author}{\bibinfo{person}{Yandong Wen}, \bibinfo{person}{Kaipeng
  Zhang}, \bibinfo{person}{Zhifeng Li}, {and} \bibinfo{person}{Yu Qiao}.}
  \bibinfo{year}{2016}\natexlab{}.
\newblock \showarticletitle{A discriminative feature learning approach for deep
  face recognition}. In \bibinfo{booktitle}{\emph{ECCV}}.
  \bibinfo{pages}{499--515}.
\newblock


\bibitem[Wu et~al\mbox{.}(2018)]%
        {wu2018privacy}
\bibfield{author}{\bibinfo{person}{Yifan Wu}, \bibinfo{person}{Fan Yang}, {and}
  \bibinfo{person}{Haibin Ling}.} \bibinfo{year}{2018}\natexlab{}.
\newblock \showarticletitle{Privacy-protective-gan for face de-identification}.
\newblock \bibinfo{journal}{\emph{arXiv preprint arXiv:1806.08906}}
  (\bibinfo{year}{2018}).
\newblock


\bibitem[Yang et~al\mbox{.}(2021)]%
        {yang2021study}
\bibfield{author}{\bibinfo{person}{Kaiyu Yang}, \bibinfo{person}{Jacqueline
  Yau}, \bibinfo{person}{Li Fei-Fei}, \bibinfo{person}{Jia Deng}, {and}
  \bibinfo{person}{Olga Russakovsky}.} \bibinfo{year}{2021}\natexlab{}.
\newblock \showarticletitle{A study of face obfuscation in imagenet}.
\newblock \bibinfo{journal}{\emph{arXiv preprint arXiv:2103.06191}}
  (\bibinfo{year}{2021}).
\newblock


\bibitem[Ye et~al\mbox{.}(2021a)]%
        {ye2021dynamic}
\bibfield{author}{\bibinfo{person}{Mang Ye}, \bibinfo{person}{Cuiqun Chen},
  \bibinfo{person}{Jianbing Shen}, {and} \bibinfo{person}{Ling Shao}.}
  \bibinfo{year}{2021}\natexlab{a}.
\newblock \showarticletitle{Dynamic tri-level relation mining with attentive
  graph for visible infrared re-identification}.
\newblock \bibinfo{journal}{\emph{IEEE TIFS}}  \bibinfo{volume}{17}
  (\bibinfo{year}{2021}), \bibinfo{pages}{386--398}.
\newblock


\bibitem[Ye et~al\mbox{.}(2021b)]%
        {ye2021collaborative}
\bibfield{author}{\bibinfo{person}{Mang Ye}, \bibinfo{person}{He Li},
  \bibinfo{person}{Bo Du}, \bibinfo{person}{Jianbing Shen},
  \bibinfo{person}{Ling Shao}, {and} \bibinfo{person}{Steven~CH Hoi}.}
  \bibinfo{year}{2021}\natexlab{b}.
\newblock \showarticletitle{Collaborative refining for person re-identification
  with label noise}.
\newblock \bibinfo{journal}{\emph{IEEE TIP}}  \bibinfo{volume}{31}
  (\bibinfo{year}{2021}), \bibinfo{pages}{379--391}.
\newblock


\bibitem[Ye et~al\mbox{.}(2021c)]%
        {ye2021deep}
\bibfield{author}{\bibinfo{person}{Mang Ye}, \bibinfo{person}{Jianbing Shen},
  \bibinfo{person}{Gaojie Lin}, \bibinfo{person}{Tao Xiang},
  \bibinfo{person}{Ling Shao}, {and} \bibinfo{person}{Steven~CH Hoi}.}
  \bibinfo{year}{2021}\natexlab{c}.
\newblock \showarticletitle{Deep learning for person re-identification: A
  survey and outlook}.
\newblock \bibinfo{journal}{\emph{IEEE TPAMI}} (\bibinfo{year}{2021}).
\newblock


\bibitem[Ye et~al\mbox{.}(2022)]%
        {ye2022aug}
\bibfield{author}{\bibinfo{person}{Mang Ye}, \bibinfo{person}{Jianbing Shen},
  \bibinfo{person}{Xu Zhang}, \bibinfo{person}{Pong~C. Yuen}, {and}
  \bibinfo{person}{Shih-Fu Chang}.} \bibinfo{year}{2022}\natexlab{}.
\newblock \showarticletitle{Augmentation Invariant and Instance Spreading
  Feature for Softmax Embedding}.
\newblock \bibinfo{journal}{\emph{IEEE TAPMI}} \bibinfo{volume}{44},
  \bibinfo{number}{2} (\bibinfo{year}{2022}), \bibinfo{pages}{924--939}.
\newblock


\bibitem[You et~al\mbox{.}(2021)]%
        {you2021reversible}
\bibfield{author}{\bibinfo{person}{Zhengxin You}, \bibinfo{person}{Sheng Li},
  \bibinfo{person}{Zhenxing Qian}, {and} \bibinfo{person}{Xinpeng Zhang}.}
  \bibinfo{year}{2021}\natexlab{}.
\newblock \showarticletitle{Reversible Privacy-Preserving Recognition}. In
  \bibinfo{booktitle}{\emph{ICME}}. \bibinfo{pages}{1--6}.
\newblock


\bibitem[Zheng et~al\mbox{.}(2015)]%
        {zheng2015scalable}
\bibfield{author}{\bibinfo{person}{Liang Zheng}, \bibinfo{person}{Liyue Shen},
  \bibinfo{person}{Lu Tian}, \bibinfo{person}{Shengjin Wang},
  \bibinfo{person}{Jingdong Wang}, {and} \bibinfo{person}{Qi Tian}.}
  \bibinfo{year}{2015}\natexlab{}.
\newblock \showarticletitle{Scalable person re-identification: A benchmark}. In
  \bibinfo{booktitle}{\emph{ICCV}}. \bibinfo{pages}{1116--1124}.
\newblock


\bibitem[Zhou et~al\mbox{.}(2021a)]%
        {zhou2021domainsurvey}
\bibfield{author}{\bibinfo{person}{Kaiyang Zhou}, \bibinfo{person}{Ziwei Liu},
  \bibinfo{person}{Yu Qiao}, \bibinfo{person}{Tao Xiang}, {and}
  \bibinfo{person}{Chen~Change Loy}.} \bibinfo{year}{2021}\natexlab{a}.
\newblock \showarticletitle{Domain generalization: A survey}.
\newblock  (\bibinfo{year}{2021}).
\newblock


\bibitem[Zhou et~al\mbox{.}(2019)]%
        {zhou2019omni}
\bibfield{author}{\bibinfo{person}{Kaiyang Zhou}, \bibinfo{person}{Yongxin
  Yang}, \bibinfo{person}{Andrea Cavallaro}, {and} \bibinfo{person}{Tao
  Xiang}.} \bibinfo{year}{2019}\natexlab{}.
\newblock \showarticletitle{Omni-scale feature learning for person
  re-identification}. In \bibinfo{booktitle}{\emph{CVPR}}.
  \bibinfo{pages}{3702--3712}.
\newblock


\bibitem[Zhou et~al\mbox{.}(2021b)]%
        {zhou2021learning}
\bibfield{author}{\bibinfo{person}{Kaiyang Zhou}, \bibinfo{person}{Yongxin
  Yang}, \bibinfo{person}{Andrea Cavallaro}, {and} \bibinfo{person}{Tao
  Xiang}.} \bibinfo{year}{2021}\natexlab{b}.
\newblock \showarticletitle{Learning generalisable omni-scale representations
  for person re-identification}.
\newblock \bibinfo{journal}{\emph{IEEE TPAMI}} (\bibinfo{year}{2021}).
\newblock


\bibitem[Zhou et~al\mbox{.}(2021c)]%
        {zhou2021domain}
\bibfield{author}{\bibinfo{person}{Kaiyang Zhou}, \bibinfo{person}{Yongxin
  Yang}, \bibinfo{person}{Yu Qiao}, {and} \bibinfo{person}{Tao Xiang}.}
  \bibinfo{year}{2021}\natexlab{c}.
\newblock \showarticletitle{Domain generalization with mixstyle}.
\newblock \bibinfo{journal}{\emph{arXiv preprint arXiv:2104.02008}}
  (\bibinfo{year}{2021}).
\newblock


\bibitem[Zhu et~al\mbox{.}(2017)]%
        {zhu2017unpaired}
\bibfield{author}{\bibinfo{person}{Jun-Yan Zhu}, \bibinfo{person}{Taesung
  Park}, \bibinfo{person}{Phillip Isola}, {and} \bibinfo{person}{Alexei~A
  Efros}.} \bibinfo{year}{2017}\natexlab{}.
\newblock \showarticletitle{Unpaired image-to-image translation using
  cycle-consistent adversarial networks}. In \bibinfo{booktitle}{\emph{ICCV}}.
  \bibinfo{pages}{2223--2232}.
\newblock


\end{thebibliography}

\end{document}